\documentclass[%
 reprint,
superscriptaddress,
 amsmath,amssymb,
 aps,
]{revtex4-2}

\usepackage{graphicx}
\usepackage{dcolumn}
\usepackage{bm}
\usepackage{mathtools}
\usepackage{multirow}
\usepackage[ruled,vlined]{algorithm2e}
\usepackage{hyperref}

\usepackage[normalem]{ulem}

\hypersetup{
  colorlinks   = true, 
  urlcolor     = black, 
  linkcolor    = blue, 
  citecolor   = magenta 
}

\newcommand*{\myprime}{\scalebox{0.8}{\scriptsize$\prime$}}
\newcommand*{\mydprime}{\scalebox{0.8}{\scriptsize$\prime$\scriptsize$\prime$}}

\begin{document}

\preprint{APS/123-QED}

\title{Diffusion probabilistic models enhance variational autoencoder for crystal structure generative modeling}

\author{Teerachote Pakornchote}
\author{Natthaphon Choomphon-anomakhun}
\author{Sorrjit Arrerut}
\affiliation{%
Extreme Conditions Physics Research Laboratory (ECPRL) and Center of Excellence in Physics of Energy Materials (CE:PEM), Department of Physics, Faculty of Science, Chulalongkorn University, Bangkok, 10330, Thailand
}

\author{Chayanon Atthapak}
\author{Sakarn Khamkaeo}
\affiliation{%
Extreme Conditions Physics Research Laboratory (ECPRL) and Center of Excellence in Physics of Energy Materials (CE:PEM), Department of Physics, Faculty of Science, Chulalongkorn University, Bangkok, 10330, Thailand
}
\affiliation{%
Thailand Center of Excellence in Physics , Ministry of Higher Education, Science, Research and Innovation, 328 Si Ayutthaya Road, Bangkok 10400, Thailand
}

\author{Thiparat Chotibut}
\affiliation{%
 Chula Intelligent and Complex Systems, Department of Physics, Faculty of Science, Chulalongkorn University, Bangkok, Thailand, 10330
}

\author{Thiti Bovornratanaraks}
\email[Correspondence to: ]{thiti.b@chula.ac.th}
\affiliation{%
Extreme Conditions Physics Research Laboratory (ECPRL) and Center of Excellence in Physics of Energy Materials (CE:PEM), Department of Physics, Faculty of Science, Chulalongkorn University, Bangkok, 10330, Thailand
}
\affiliation{%
Thailand Center of Excellence in Physics , Ministry of Higher Education, Science, Research and Innovation, 328 Si Ayutthaya Road, Bangkok 10400, Thailand
}

\date{\today}

\begin{abstract}
The crystal diffusion variational autoencoder (CDVAE) is a machine learning model that leverages score matching to generate realistic crystal structures that preserve crystal symmetry. In this study, we leverage novel diffusion probabilistic (DP) models to denoise atomic coordinates rather than adopting the standard score matching approach in CDVAE. Our proposed DP-CDVAE model can reconstruct and generate crystal structures whose qualities are statistically comparable to those of the original CDVAE. Furthermore, notably, when comparing the carbon structures generated by the DP-CDVAE model with relaxed structures obtained from density functional theory calculations, we find that the DP-CDVAE generated structures are remarkably closer to their respective ground states. The energy differences between these structures and the true ground states are, on average, 68.1 meV/atom lower than those generated by the original CDVAE. This significant improvement in the energy accuracy highlights the effectiveness of the DP-CDVAE model in generating crystal structures that better represent their ground-state configurations.

\begin{description}
\item[Keywords]
denoising diffusion probabilistic models, variational autoencoder, \\crystal structures, ground state, density functional theory
\item[Open data] The dataset used in this work has been made available at \\ \url{https://github.com/trachote/dp-cdvae}
\end{description}
\end{abstract}

\maketitle


\section{Introduction}
Advances in computational materials science has enabled the accurate prediction of novel materials possessing exceptional properties. Remarkably, these computational advancements have facilitated the successful experimental synthesis of materials that exhibit the anticipated properties. Some predicted materials, such as near-room-temperature superconductors, have been successfully synthesized under high-pressure conditions, with their superconducting temperatures in accordance with density functional theory (DFT) calculations \cite{Needs2016, Kohn1965}. To achieve accurate predictions, \textit{a priori} knowledge of plausible molecular and crystal structures play a vital role in both theoretical and experimental studies. Several algorithms, such as evolutionary algorithms, swarm particle optimization, random sampling method, and etc., have been employed for structure prediction \cite{Uspex2006, Calypso2010, Pickard2011}. These algorithms rely on identifying local minima on the potential energy landscape obtained from DFT calculations \cite{Oganov2019-dw, Schon2010}. In the case of crystal structures, where atoms are arranged in three-dimensional space with periodic boundaries, additional criteria are necessary to enforce crystal symmetry constraints \cite{Pickard2011}. 

Recent approach for structure prediction employs denoising diffusion models to perform probabilistic inference. These models sample molecular and crystal structures from a probability distribution of atomic coordinates and types \cite{Xie2022cdvae, Shi2021, Xu2022geodiff, Guan2023edm}, bypassing the computationally intensive DFT calculation to tediously determine the potential energy landscape. By leveraging sufficiently large datasets containing various compounds, this method enables the generation of diverse compositions and combinations of elements simultaneously. Furthermore, the models allow for the control of desired physical properties of the generated structures through conditional probability sampling \cite{Kang2019, Lim2018-mt, Song2022solving}. These machine learning-based algorithms also hold promise for solving inverse problem efficiently, resolving structures from experimental characterizations, e.g., x-ray absorption spectroscopy and other techniques, a challenging problem in materials science \cite{Cui2019, Carbone2020, Liang2023}.

There are two primary types of denoising diffusion models: score matching approach and denoising diffusion probabilistic models (DDPM) \cite{Song2019, Song2021, Ho2020}. These two models can denoise (reverse) a normal distribution such that the distribution gradually transforms into the data distribution of interest.  The score matching approach estimates the score function of the perturbed data directing the normal distribution toward the data distribution and employing large step sizes of variance. In contrast, DDPM gradually denoises the random noise through a joint distribution of data perturbed at different scales of variance. 



Since atomic positions in crystal structures are periodic and can be invariant under some rotation groups depending on their crystal symmetry, the core neural networks should favourably possess roto-translational equivariance \cite{Bronstein2021, Cohen2018, Thomas2018}. Xie et al. \cite{Xie2022cdvae} has proposed a model for crystal prediction by a combination between variational autoencoder (VAE) \cite{Kingma2014} and the denoising diffusion model, called crystal diffusion VAE (CDVAE). The model employs the score matching approach with (annealed) Langevin dynamics to generate new crystal structures from random coordinates \cite{Song2019}.  
The neural networks for an encoder and the diffusion model are roto-translationally equivariant. As a result, CDVAE can generate crystal structures with realistic bond lengths and respect crystal symmetry. 

Because of the periodic boundary condition imposed on the unit cell, gradually injecting sufficiently strong noises (in the forward process) to the  fractional coordinates can lead to the uniform distribution of atomic positions at late times, the consequence of ergodicity in statistical mechanical sense. Rather than beginning with a Gaussian distribution and denoising it as in the original CDVAE formulation, Jiao et al. \cite{Jiao2023diffcsp} perturbed and sampled atomic positions beginning with a wrapped normal distribution which satisfies the periodic boundary condition. With this approach, the reconstruction performance has been significantly improved.  Other circular (periodic) distributions, e.g., the wrapped normal and von Mises distributions, are not natural for DDPM framework since there is no known analytical method to explicitly incorporate such distributions into the framework. There, one needs to resort to an additional sampling procedure to construct the DDPM \cite{Okhotin2023starshaped}.

In this work, we introduce a crystal generation framework called diffusion probabilistic CDVAE (DP-CDVAE). Similar to the original CDVAE, our model consists of two parts: the VAE part and the diffusion part. The purpose of the VAE part is to predict the lattice parameters and the number of atoms in the unit cell of crystal structures. On the other hand, the diffusion part utilizes the diffusion probabilistic approach to denoise fractional coordinates and predict atomic coordinates. By employing the DDPM instead of the score matching approach, the DP-CDVAE model shows reconstruction and generation task performances that are statistically comparable to those obtained from original CDVAE. Importantly, we demonstrate the significantly higher ground-state generation performance of DP-CDVAE, through the distance comparison between generated structures and those optimized using the DFT method. We also analyze the changes in energy and volume after relaxation to gain further insights into models' capabilities.


\begin{figure*}[t]
\includegraphics[width=1.\textwidth]{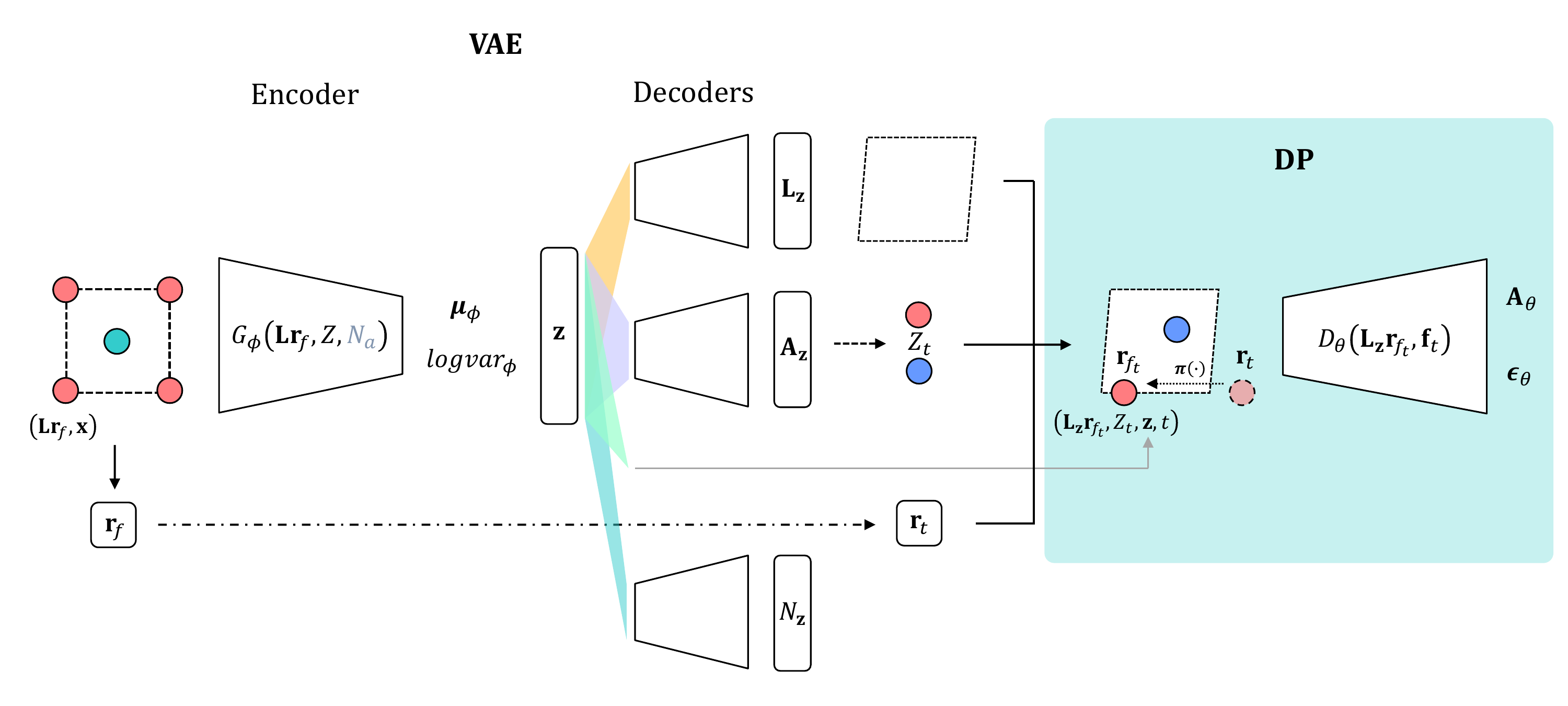}
\caption{\label{fig:model}The schematic summarizing the architecture for training the DP-CDVAE model. Multiple sub-networks are trained to minimize the total loss function of Eq.\eqref{eq:total_loss}. The encoder ($G_{\phi}(\mathbf{L}\mathbf{r}_f, Z, N_a)$) compresses input pristine crystal structures into the latent feature ($\mathbf{z}$). The predicted lattice parameters ($\mathbf{L}_\mathbf{z}$), the predicted number of atoms ($N_\mathbf{z}$), and $\mathbf{A}_\mathbf{z}$ are decoded from $\mathbf{z}$. Here, $\mathbf{A}_\mathbf{z}$ enables the sampling of atomic types ($Z_t$), and all the decoded features enable the reconstruction of crystal structures. The input fractional coordinates $\mathbf{r}_f$ undergo perturbation (dash-dotted line) at time step $t$ and then are transformed by $\boldsymbol{\pi}(\cdot)$ to satisfy the periodic boundary condition (dotted line), serving as the coordinates for the reconstructed crystal structures. These reconstructed structures, $\left( \mathbf{L_z r}_{f_t}, Z_t, \mathbf{z},t \right)$, are subsequently fed into the diffusion network ($D_{\theta}(\mathbf{L}_\mathbf{z}\mathbf{r}_{f_t}, \mathbf{f}_{t})$), where $\mathbf{f}_t$ is a node feature composing of $Z_t$, $\mathbf{z}$, and $t$. The diffusion network predicts the noise added to the fractional coordinates ($\boldsymbol{\epsilon}_{\theta}$) as well as the one-hot vector of atomic types ($\mathbf{A}_{\theta}$), see Sec. \ref{sec: DP-CDVAE Arch}. Dashed-line boxes represent the unit cells of the crystal structures.
}
\end{figure*}

\section{Results}
The performances of DP-CDVAE models are herein presented. There are four DP-CDVAE models, differed by the choice of the encoder (see Fig.~\ref{fig:encoders}). DimeNet$^{++}$ has been employed for the main encoder for every DP-CDVAE models \cite{Gasteiger2022dimenetpp}. We then modify the encoder of DP-CDVAE to encode the crystal structure by two additional neural networks: a multilayer perceptron that takes the number of atoms in the unit cell ($N_a$) as an input, and a graph isomorphism network (GINE) \cite{Hu2020gine}. Their latent features are combined with the latent features from DimeNet$^{++}$ through another multilayer perceptron. The $N_a$ is encoded such that the model can decode the $N_a$ accurately, and GINE encoder is inspired by GeoDiff \cite{Xu2022geodiff} whose model is a combination of SchNet \cite{Schutt2017schnet} and GINE which yields better performance. 

Three datasets, \textbf{Perov-5} \cite{Castelli2012perov5, Castelli2012perov5-2}, \textbf{Carbon-24} \cite{Carbon2020data}, and \textbf{MP-20} \cite{Jain2013mp20}, were selected to evaluate the performance of the model. The Perov-5 dataset consists of perovskite materials with cubic structures, but with variations in the combinations of elements within the structures. The Carbon-24 dataset comprises carbon materials, where the data consists of carbon element with various crystal systems obtained from \textit{ab initio} random structure searching algorithm at pressure of 10 GPa \cite{Carbon2020data}. The MP-20 dataset encompasses a wide range of compounds and structure types. 

\subsection{Reconstruction performance}
The reconstruction performance is determined by the similarity between reconstructed and ground-truth structures. The similarity can be evaluated using Niggli's algorithm implemented in \texttt{StructureMatcher} method from \texttt{pymatgen} library \cite{Grosse-Kunstleve2004}. The reconstructed and ground-truth structures are similar if they pass the criteria of \texttt{StructureMatcher} which are \texttt{stol=0.5, angle\_tol=10, ltol=0.3}. \textit{Match rate} is the percentage of those structures passed the 
criteria. If the reconstructed and ground-truth structures are similar under the criteria, the root-mean-square distance between their atomic positions is computed and then normalized by $\sqrt[3]{V/N_a}$, where $V$ is the unit-cell volume, and $N_a$ is the number of atoms in the unit cell. An average of the distances of every pair of structures ($\langle\delta_{\text{rms}}\rangle
$), computed from Niggli's algorithm, is used as the performance metric.

Table~\ref{tab:reconstruction} presents the reconstruction performance of different models for three different datasets: Perov-5, Carbon-24, and MP-20. For the Perov-5 dataset, the DP-CDVAE model achieves a match rate of 90.04\%, indicating its ability to reconstruct a significant portion of the ground-truth structures. This performance is slightly lower than the CDVAE model but still demonstrates the effectiveness of our model. In terms of $\langle\delta_{\text{rms}}\rangle
$, the DP-CDVAE model achieves a value of 0.0212, comparable to the FTCP model \cite{Ren2022}, but slightly higher than the CDVAE model. Similarly, for the Carbon-24 and MP-20 datasets, the DP-CDVAE model performs well in terms of both match rate and $\langle\delta_{\text{rms}}\rangle
$. It achieves match rates of 45.57\% and 32.42\% for Carbon-24 and MP-20, respectively. The corresponding $\langle\delta_{\text{rms}}\rangle
$ values for Carbon-24 and MP-20 are 0.1513 and 0.0383, respectively, comparable to the CDVAE model.

Regarding the DP-CDVAE+$N_a$ model, the additional encoding of $N_a$ into the model leads to improved match rates for all datasets, with an increase of 2--5\%. This enhancement can be attributed to the accurate prediction of $N_a$. However, in terms of $\langle\delta_{\text{rms}}\rangle$, only the Perov-5 dataset shows an improvement, with a value of 0.0149. On the other hand, for the Carbon-24 and MP-20 datasets, the $\langle\delta_{\text{rms}}\rangle$ values are higher compared to the DP-CDVAE model.

For the DP-CDVAE+GINE and DP-CDVAE+$N_a$+GINE models, the additional encoding of GINE into the models leads to a substantial drop in match rates compared to the DP-CDVAE model, particularly for the Perov-5 and Carbon-24 datasets. In contrast, there is a moderate increase in the match rates for the MP-20 dataset. The $\langle\delta_{\text{rms}}\rangle$ values for the Perov-5 and Carbon-24 datasets are comparable to those of the DP-CDVAE model. However, for the MP-20 dataset, the $\langle\delta_{\text{rms}}\rangle$ is noticeably higher in the models with GINE encoder compared to the DP-CDVAE model.

Overall, while the reconstruction performance of the DP-CDVAE model may be slightly lower than the CDVAE model in terms of match rate, it still demonstrates competitive performance with relatively low $\langle\delta_{\text{rms}}\rangle$. The match rate can be enhanced by additionally encoding the $N_a$, but the performance is traded off by the increase in $\langle\delta_{\text{rms}}\rangle$.

\begin{table*}
\caption{Reconstruction performance.}
\label{tab:reconstruction}
\begin{ruledtabular}
\begin{tabular}{lcccccc}
\multicolumn{1}{l}{Models} & \multicolumn{3}{c}{Match rate (\%) $\uparrow$} & \multicolumn{3}{c}{$\langle\delta_{\text{rms}}\rangle$ $\downarrow$} \\
\cline{2-4} \cline{5-7}
 & Perov-5 & Carbon-24 & MP-20 & Perov-5 & Carbon-24 & MP-20 \\
\hline
FTCP~\cite{Xie2022cdvae} & \textbf{99.34} & \textbf{62.28} & \textbf{69.89} & 0.0259 & 0.2563 & 0.1593 \\
CDVAE~\cite{Xie2022cdvae} & 97.52 & 55.22 & 45.43 & 0.0156 & \textbf{0.1251} & \textbf{0.0356} \\
DP-CDVAE & 90.04 & 45.57 & 32.42 & 0.0212 & 0.1513 & 0.0383 \\
DP-CDVAE+$N_a$ & 91.86 & 50.99 & 36.17 & \textbf{0.0149} & 0.1612 & 0.0560 \\
DP-CDVAE+GINE & 80.50 & 49.02 & 34.08 & 0.0214 & 0.1599 & 0.0455 \\
DP-CDVAE+$N_a$+GINE & 88.30 & 38.28 & 37.44 & 0.0180 & 0.1921 & 0.0525 \\

\end{tabular}
\end{ruledtabular}
\end{table*}

\subsection{Generation performance}
We follow the CDVAE model that used three metrics to determine the generation performance of the models  \cite{Xie2022cdvae}. The first metric is \textit{Validity} percentage. The structures that are valid for structure and composition tests must satisfy two criteria: distances of every pair of atoms are larger than 0.5 {\AA} and the total charge in the unit cell is neutral. The second metric is called {\it coverage} (COV), which utilizes structure and composition fingerprints to evaluate the similarity between the generated and ground-truth structures. COV-R (Recall) represents the percentage of ground-truth structures covered by the generated structures. COV-P (Precision) represents the percentage of generated structures that are similar to the ground-truth structures, indicating the quality of the generation.  The third metric is the Wasserstein distance between property distributions of generated and ground-truth structures. Three property statistics are density ($\rho$), which is total atomic mass per volume (unit g/cm$^3$), formation energy ($E_{form}$, unit eV/atom), and the number of elements in the unit cell (\# elem.). A separated and pre-trained neural network is employed to predict $E$ of the structures where the detail of the pre-training can be found in 
Ref.~\cite{Xie2022cdvae}. The first and second metrics are computed over 10,240 generated structures, and 1000 structures are randomly chosen from the generated structures that pass the validity tests to compute the third metric. The ground-truth structures used to evaluate the generation performance are from the test set.

In Table~\ref{tab:generation}, the DP-CDVAE model achieves a validity rate of 100\% for the Perov-5 dataset and close to 100\% for the Carbon-24 and MP-20 datasets in terms of structure. The validity rate for composition is comparable to that of the CDVAE model. The DP-CDVAE model also demonstrates comparable COV-R values to the CDVAE model across all three datasets. Furthermore, the DP-CDVAE models with $N_a$ and/or GINE encoders exhibit similar Validity and COV-R metrics to those of the DP-CDVAE model. However, for COV-P, all DP-CDVAE models yield lower values compared to CDVAE.

On the other hand, our models show significant improvements in property statistics. In the case of the MP-20 dataset, the DP-CDVAE models, particularly those with the GINE encoder, yield substantially smaller Wasserstein distances for $\rho$, $E_{form}$, and the number of elements compared to other models. For the Carbon-24 dataset, our models also exhibit a smaller Wasserstein distance for $\rho$ compared to the CDVAE model.


\subsection{Ground-state performance}
Another objective of the structure generator is to generate novel structures that also are close to the ground state. To verify that, the generated structures are relaxed using the DFT calculation where the \textit{relaxed structures} exhibit balanced internal stresses with external pressures and reside in local energy minima. These relaxed structures are then compared with the generated structures to evaluate their similarity. In this study, we have chosen a set of 100 generated structures from each of CDVAE, CDVAE+Fourier, and DP-CDVAE models for relaxation where CDVAE+Fourier model is CDVAE model with Fourier embedding features of the perturbed coordinates. However, relaxation procedures for multi-element compounds can be computationally intensive. To address this, we have specifically selected materials composed solely of carbon atoms, using the model trained on Carbon-24 dataset. This selection ensures a convergence of the self-consistent field in DFT calculation. Moreover, in the relaxation, we consider the ground state of the relaxed structures at a temperature of 0 K and a pressure of 10 GPa since the carbon structures in the training set are stable at 10 GPa \cite{Carbon2020data}.

We here introduce a ground-state performance presented in Table~\ref{tab:ground-state}. The \texttt{StructureMatcher} with the same criteria as in the reconstruction performance is used to evaluate the similarity between the generated and relaxed structures. The relaxed structure was used as a based structure to determine if the generated structure can be matched. Four metrics used to determine the similarity are 1) match rate, 2) $\langle\delta_{\text{rms}}\rangle$, 3) $\Delta V_{\text{rms}}$ and 4) $\Delta E_{\text{rms}}$. 
The $\Delta V_{\text{rms}}$ and $\Delta E_{\text{rms}}$ represent the root mean square differences in volume and energy, respectively, between the generated structures and the relaxed structures in the dataset.

In Table~\ref{tab:ground-state}, the DP-CDVAE model achieves the highest match rate and the lowest $\langle\delta_{\text{rms}}\rangle$ and $\Delta E_{\text{rms}}$. Although the CDVAE+Fourier model achieves the lowest $\Delta V_{\text{rms}}$, the DP-CDVAE model demonstrates the $\Delta V_{\text{rms}}$ that is comparable to that of the CDVAE+Fourier model. 

\begin{table*}
\caption{Generation performance}
\label{tab:generation}
\begin{ruledtabular}
\begin{tabular}{llccccccc}
\multirow{2}{*}{Datasets} & \multirow{2}{*}{Models} & \multicolumn{2}{c}{Validity (\%) $\uparrow$} & \multicolumn{2}{c}{COV (\%) $\uparrow$} & \multicolumn{3}{c}{Property statistics $\downarrow$} \\
\cline{3-4} \cline{5-6} \cline{7-9}
& & Struc. & Comp. & R. & P. & $\rho$ & $E_{form}$ & \# elem. \\
\hline
\multirow{7}{*}{Perov-5} & G-SchNet~\cite{Xie2022cdvae} & 99.92 & 98.79 & 0.18 & 0.23 & 1.625 & 4.746 & 0.0368 \\
& P-G-SchNet~\cite{Xie2022cdvae} & 79.63 & \textbf{99.13} & 0.37 & 0.25 & 0.2755 & 1.388 & 0.4552 \\
& CDVAE~\cite{Xie2022cdvae} & \textbf{100} & 98.59 & 99.45 & \textbf{98.46} & 0.1258 & \textbf{0.0264} & 0.0628 \\
& DP-CDVAE & \textbf{100} & 98.07 & 99.52 & 98.39 & 0.1807 & 0.0713 & 0.0767 \\
& DP-CDVAE+$N_a$ & 99.99 & 97.34 & \textbf{99.55} & 97.22 & \textbf{0.1027} & 0.0287 & 0.0437 \\
& DP-CDVAE+GINE & \textbf{100} & 96.11 & 98.94 & 95.63 & 0.2114 & 0.0832 & 0.0498 \\
& DP-CDVAE+$N_a$+GINE & \textbf{100} & 97.09 & 99.52 & 96.73 & 0.1368 & 0.0425 & \textbf{0.0210} \\
\hline
\multirow{7}{*}{Carbon-24} & G-SchNet~\cite{Xie2022cdvae} & 99.94 & -- & 0.00 & 0.00 & 0.9427 & 1.320 & -- \\
& P-G-SchNet~\cite{Xie2022cdvae} & 48.39 & -- & 0.00 & 0.00 & 1.533 & 134.7 & -- \\
& CDVAE~\cite{Xie2022cdvae} & \textbf{100} & -- & 99.80 & \textbf{83.08} & 0.1407 & 0.2850 & -- \\
& DP-CDVAE & 99.92 & -- & 99.56 & 77.98 & 0.1109 & \textbf{0.2596} & -- \\
& DP-CDVAE+$N_a$ & 99.73 & -- & 99.61 & 72.29 & 0.1080 & 0.3030 & -- \\
& DP-CDVAE+GINE & 99.50 & -- & \textbf{100} & 68.13 & \textbf{0.0977} & 0.3623 & -- \\
& DP-CDVAE+$N_a$+GINE & 98.61 & -- & 99.21 & 65.13 & 0.1267 & 0.4136 & -- \\
\hline
\multirow{7}{*}{MP-20} & G-SchNet~\cite{Xie2022cdvae} & 99.65 & 75.96 & 38.33 & 99.57 & 3.034 & 42.09 & 0.6411 \\
& P-G-SchNet~\cite{Xie2022cdvae} & 77.51 & 76.40 & 41.93 & \textbf{99.74} & 4.04 & 2.448 & 0.6234 \\
& CDVAE~\cite{Xie2022cdvae} & \textbf{100} & \textbf{86.70} & 99.15 & 99.49 & 0.6875 & 0.2778 & 1.432 \\
& DP-CDVAE & 99.59 & 85.44 & 98.93 & 98.96 & 0.4037 & 0.1547 & 0.9179 \\
& DP-CDVAE+$N_a$ & 99.81 & 84.95 & 99.36 & 99.33 & 0.4889 & 0.1800 & 1.053 \\
& DP-CDVAE+GINE & 99.82 & 81.92 & \textbf{99.48} & 99.00 & 0.2785 & 0.0603 & \textbf{0.5679} \\
& DP-CDVAE+$N_a$+GINE & 99.90 & 83.89 & 95.51 & 99.27 & \textbf{0.1790} & \textbf{0.0522} & 0.6909
\end{tabular}
\end{ruledtabular}
\end{table*}

\begin{table*}
\caption{\label{tab:ground-state}Ground-state performance.}
\begin{ruledtabular}
\begin{tabular}{ccccc}
     Model & Match rate (\%) $\uparrow$ & $\langle\delta_{\text{rms}}\rangle$ $\downarrow$ & $\Delta V_{\text{rms}}$ ({\AA}$^3$/atom) $\downarrow$ & $\Delta E_{\text{rms}}$ (meV/atom) $\downarrow$ \\
     \hline
     CDVAE & 63 & 0.0321 & 0.0227 & 468.8 \\
     CDVAE+Fourier & 62 & 0.0216 & \textbf{0.0157} & 494.4 \\
     DP-CDVAE & \textbf{64} & \textbf{0.0141} & 0.0158 & \textbf{400.7} \\
\end{tabular}
\end{ruledtabular}
\end{table*}

\section{Discussion}
The DP-CDVAE models significantly enhance the generation performance, particularly in terms of property statistics, while maintaining comparable COVs to those of CDVAE. Specifically, for Carbon-24 and MP-20 datasets, the density distributions between the generated and ground-truth structures from DP-CDVAE models exhibit substantially smaller Wasserstein distance compared those of the CDVAE model (see Table~\ref{tab:generation}). The $\Delta V_{\text{rms}}$ of the DP-CDVAE model presented in Table~\ref{tab:ground-state} is significantly lower than that of the original CDVAE. This is corresponding to smaller Wasserstein distance of $\rho$ shown in Table~\ref{tab:generation}. The DP-CDVAE model also demonstrates significantly smaller $\langle\delta_{\text{rms}}\rangle$ than the original CDVAE. These suggest that our lattice generation closely approximates the relaxed lattice, while also achieving atomic positions that closely resemble the ground-state configuration. Additionally, the distribution of the number of elements in the unit cells is relatively similar to that of the data in the test set, particularly in the results from the models with GINE encoder. This could be attributed to the capability of GINE to search for graph isomorphism \cite{Xu2018gin}.

Moreover, $\Delta E$ is the energy difference between the generated structures and their corresponding relaxed structures. The ground-state energy represents a local minimum that the generated structure is relaxed towards. A value of $\Delta E$ close to zero indicates that the generated structure is in close proximity to the ground state. 
In Table~\ref{tab:ground-state}, it can be observed that our model achieves the $\Delta E_{\text{rms}}$ value of 400.7 meV/atom which is about 68.1 meV/atom lower than the $\Delta E_{\text{rms}}$ of CDVAE. 
The mode of $\Delta E$ of our model is 64 -- 128 meV/atom, which is lower than its root-mean-square value (see Fig.~\ref{fig:histograms}). 
Nevertheless, both the $\Delta E_{\text{rms}}$ and the mode of $\Delta E$ exhibit relatively high values. In many cases, the formation energy of synthesized compounds is reported to be above the convex hull less than 36 meV/atom \cite{Wu2013, Ishikawa2020, Ektarawong2023}. To obviate the need for time-consuming DFT relaxation, it is essential for the generated structures to be even closer to the ground state. Therefore, achieving lower $\Delta E_{\text{rms}}$ values remains a milestone for future work.


\section{Methods}
\subsection{Diffusion probabilistic model}
In the diffusion probabilistic model, the data distribution is gradually perturbed by noise in the forward process until it becomes a normal distribution at late times. In this study, the distribution of the fractional coordinate ($\mathbf{r}_f$) is considered since their values of every crystal structures distribute over the same range ,i.e., $\mathbf{r}_f \in [0, 1)^3$. The Markov process is assumed for the forward diffusion such that the joint distribution is a product of the conditional distributions conditioned on the knowledge of the fractional coordinate at the previous time step:
\begin{equation}
\begin{aligned}
   q(\mathbf{r}_{1:T}|\mathbf{r}_0) &= \prod_{t=1}^T q(\mathbf{r}_t|\mathbf{r}_{t-1}),
\\
    q(\mathbf{r}_t|\mathbf{r}_{t-1}) &= \mathcal{N}(\mathbf{r}_t; \sqrt{\alpha_t}\mathbf{r}_{t-1}, (1-\alpha_t)\mathbf{I}),  
\end{aligned}   
\end{equation}
where $\mathbf{r}_0 \sim q(\mathbf{r}_f)$ which is the data distribution of the fractional coordinate, $t$ is the discretized diffusion time step, $T$ is the final diffusion time, $\alpha_t$ is a noise schedule with a sigmoid scheduler \cite{Kingma2021vdm}, and the conditional $q(\cdot | \cdot)$ is a Gaussian kernel due to the Markov diffusion process assumption. Then $\mathbf{r}_t$ can be expressed in the Langevin's form through the reparameterization trick as
\begin{equation}
\label{eq:r_t}
\mathbf{r}_t = \sqrt{\bar{\alpha}_t}\mathbf{r}_0 + \sqrt{1 - \bar{\alpha}_t}\boldsymbol{\epsilon},
\end{equation}
where $\boldsymbol{\epsilon} \sim \mathcal{N}(0,\mathbf{I})$, and $\bar{\alpha}_t = \prod_{i=1}^t\alpha_i$. This update rule does not necessitate $\mathbf{r}_t$ to remain in $[0,1)^3$; however, we can impose the periodic boundary condition for the fractional coordinate so that 
\begin{equation}
    \mathbf{r}_{f_t} = \boldsymbol{\pi}(\mathbf{r}_t) \coloneqq \mathbf{r}_t - \lfloor \mathbf{r}_t \rfloor.
\end{equation}
Then, $\mathbf{r}_{f_t} \in [0, 1)^3$.

\begin{figure*}[t]
\includegraphics[width=1.\textwidth]{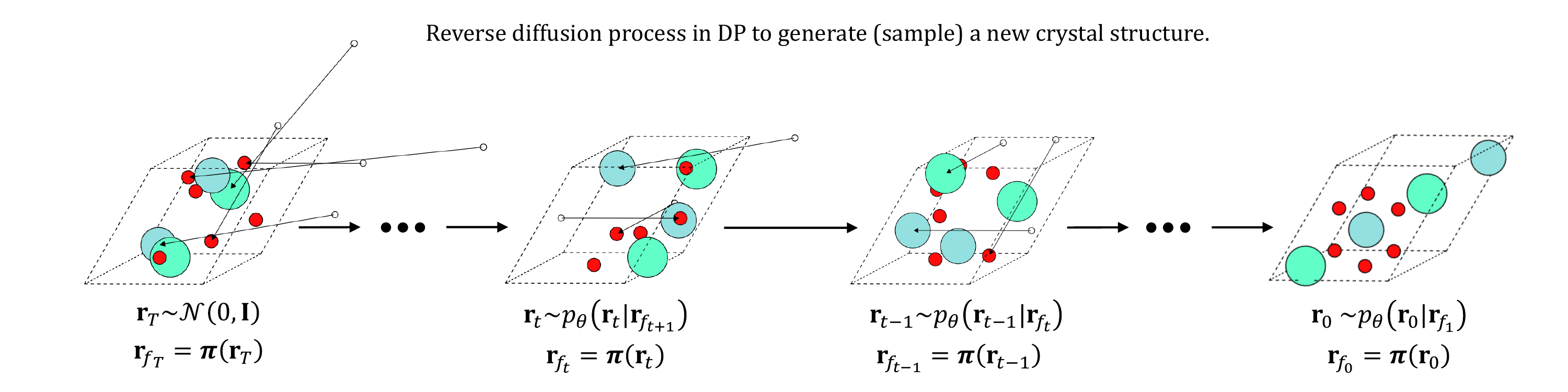}
\caption{\label{fig:reverse}
The schematic depicting the reverse diffusion process of the DP-CDVAE model.  Initially, atomic coordinates are sampled from a normal distribution and subsequently mapped into the unit cell (dashed-line box) using the periodic boundary-imposing function $\boldsymbol{\pi}(\cdot)$. White circles  outside the unit cell depict the atomic coordinates prior to the periodic boundary condition is imposed, while colored circles represent atoms that are inside the unit cell of interest. The action of $\boldsymbol{\pi}(\cdot)$ on the atoms outside the unit cell is represented by an arrow that translates the white circles into colored circles in the unit cell. Left to right show the reverse direction of the arrow of time, depicting the reverse diffusion process.}
\end{figure*}

In the reverse diffusion process, if the consecutive discretized time step is small compared to the  diffusion timescale, the reverse coordinate trajectories can be approximately sampled also from the product of Gaussian diffusion kernels as
\begin{equation}
\begin{aligned}    
    p_{\theta}(\mathbf{r}_{0:T})&= p(\mathbf{r}_T)\prod_{t=1}^Tp_{\theta}(\mathbf{r}_{t-1}|\mathbf{r}_t),
    \\
    p_{\theta}(\mathbf{r}_{t-1}|\mathbf{r}_t) &= \mathcal{N}(\mathbf{r}_{t-1};\boldsymbol{\mu}_{\theta},\sigma^2_t\mathbf{I}),
\end{aligned}
\end{equation}
where
\begin{equation}
\begin{aligned}
    \boldsymbol{\mu}_{\theta} &= \frac{1}{\sqrt{\alpha_t}}\Big(\mathbf{r}_t -\frac{1-\alpha_t}{\sqrt{1-\bar{\alpha}_t}}\boldsymbol{\epsilon}_{\theta}\Big), \\
    \sigma_t^2 &= \frac{(1 - \bar{\alpha}_{t-1})(1 - \alpha_t)}{1 - \bar{\alpha}_t}.
\end{aligned}
\end{equation}
The reverse conditional distribution can be trained by minimizing the Kullback–Leibler divergence between $p_{\theta}(\mathbf{r}_{t-1}|\mathbf{r}_t)$ and $q(\mathbf{r}_{t-1}| \mathbf{r}_t, \mathbf{r}_0)$, the posterior of the corresponding forward process \cite{Ho2020}. We use GemNetT for the diffusion network to train the parametrized noise $\boldsymbol{\epsilon}_{\theta}$  \cite{Klicpera2021gemnet}. Then, the coordinate in the earlier time can be sampled from $\mathbf{r}_{t-1} \sim p_{\theta}(\mathbf{r}_{t-1}|\mathbf{r}_t)$, whose corresponding reverse Langevin's dynamics reads
\begin{equation}
\label{eq:r_t-1}
    \mathbf{r}_{t-1} = \frac{1}{\sqrt{\alpha_t}}\Big(\mathbf{r}_t -\frac{1-\alpha_t}{\sqrt{1-\bar{\alpha}_t}}\boldsymbol{\epsilon}_{\theta}\Big) + \sigma_t\boldsymbol{\epsilon}^{\myprime},
\end{equation}
where $\boldsymbol{\epsilon}^{\myprime} \sim \mathcal{N}(0,\mathbf{I})$. Crucially, we empirically found that the final reconstruction performance is considerably improved when we impose the periodic boundary condition on the fractional coordinate at every time step such that $\mathbf{r}_{t-1} \sim p_{\theta}(\mathbf{r}_{t-1}|\mathbf{r}_{f_t})$ and $\alpha_t$ in the first term of Eq.~\eqref{eq:r_t-1} is replaced by $\bar{\alpha}_t$. Namely, in our modified reverse process, the coordinate is sampled from
\begin{equation}
\begin{aligned}
\label{eq:rf_t-1}
    \mathbf{r}_{t-1} &= \frac{1}{\sqrt{\bar{\alpha_t}}}\Big(\mathbf{r}_{f_t} -\sqrt{1-\bar{\alpha}_t}\boldsymbol{\epsilon}_{\theta}\Big) + \sigma_t\boldsymbol{\epsilon}^{\myprime},
    \\
    \mathbf{r}_{f_{t}} &= \boldsymbol{\pi}(\mathbf{r}_{t}).
\end{aligned}
\end{equation}
An illustration of denoising atomic coordinates with Eq.~\eqref{eq:rf_t-1} is demonstrated in Fig.~\ref{fig:reverse}. The model performance using Eq.~\eqref{eq:r_t-1} is shown in Table~\ref{tab:si_reconstruction}, whereas the performance using Eq.~\eqref{eq:rf_t-1} is shown in Table~\ref{tab:reconstruction}.

\subsection{Graph neural networks}
Graph neural networks architecture facilitate machine learning of crystal graphs $\mathcal{G} =(\mathcal{V},\mathcal{E})$, graph representations of crystal structures. $\mathcal{V}$ and $\mathcal{E}$ are sets of nodes and edges, respectively, defined as
\begin{eqnarray*}
    &\mathcal{V} = \{(\mathbf{f}_{n}, \mathbf{r}_{c_{n}}) \; | \; \mathbf{f}_{n} \in \mathbb{R}^{M}, \; \mathbf{r}_{c_{n}} = \mathbf{L}\mathbf{r}_{f_{n}} \in \mathbb{R}^{3} \}   , \\
    &\mathcal{E} = \{\Delta\mathbf{r}_{c_{mn}}^{(\mathbf{T})} \; | \; \Delta\mathbf{r}_{c_{mn}}^{(\mathbf{T})} = \mathbf{r}_{c_{m}} - \mathbf{r}_{c_{n}} + \mathbf{T}; \ \mathbf{r}_{c_{m}}, \mathbf{r}_{c_{n}} \in \mathbb{R}^{3} \} ,
\end{eqnarray*}
where $n$ and $m$ are indices of atoms in a crystal structure, $\mathbf{f}_n$ is a vector of $M$ features of an atom in the unit cell, $\mathbf{T}$ is a translation vector, and $\mathbf{L}$ is a lattice matrix that converts a fractional coordinate $\mathbf{r}_{f_n}$ into its atomic Cartesian coordinate $\mathbf{r}_{c_n}$. The atomic features, fractional coordinates, and atomic Cartesian coordinates of the crystal structure are vectorized (concatenated) as $\mathbf{f} = (\mathbf{f}_1, \dots, \mathbf{f}_{N_a}) \in \mathbb{R}^{N_a\times M}$, $\mathbf{r}_f = (\mathbf{r}_{f_1}, \dots, \mathbf{r}_{f_{N_a}}) \in \mathbb{R}^{N_a\times 3}$, and $\mathbf{r}_c = (\mathbf{r}_{c_1}, \dots, \mathbf{r}_{c_{N_a}}) \in \mathbb{R}^{N_a\times 3}$. Three graph neural networks implemented in this work are DimeNet$^{++}$ \cite{Gasteiger2022dimenetpp}, GINE \cite{Hu2020gine}, and GemNetT \cite{Klicpera2021gemnet}. DimeNet$^{++}$ and GINE are employed for encoders, and GemNetT is used for a diffusion network. DimeNet$^{++}$ and GemNetT, whose based architecture concerns geometry of the graphs, are rotationally equivariant. GemNetT has been devised by incorporating the polar angles between four atoms into DimeNet$^{++}$. This development grants GemNetT a higher degree of expressive power compared to DimeNet$^{++}$ \cite{Joshi2023}. Furthermore, GINE has been developed to distinguish a graph isomorphism, but not graph geometry nor the distance between nodes, which is important for our study. Thus we supplement the edge attributes into GINE with the distances between atoms, i.e. $\mathcal{E} = \{||\Delta \mathbf{r}_{c_{mn}}^{(\mathbf{T})}|| \}$.

\subsection{DP-CDVAE's architecture}\label{sec: DP-CDVAE Arch}
The forward process of DP-CDVAE model is illustrated in Fig.~\ref{fig:model}. The model is a combination of two generative models, which are VAE and diffusion probabilistic model. The pristine crystal structures consist of the fractional coordinate ($\mathbf{r}_f$), the lattice matrix ($\mathbf{L}$), ground-truth atomic type ($Z$), and the number of atoms in a unit cell ($N_a$). For crystal graphs of the encoders, their node features are $\mathbf{f}=Z$. The number of atoms in a unit cell $N_a$ is encoded through multilayer perceptron before concatenated with the latent features from other graph encoders.
They are encoded to train $\boldsymbol{\mu}_{\phi}$ and $logvar_{\phi}$ where $\phi$ is a learnable parameter of the encoders. The latent variables ($\mathbf{z}$) can be obtained by 
\begin{equation}
\label{eq:reparam}
\mathbf{z} = \boldsymbol{\mu}_{\phi} + e^{logvar_{\phi}}\boldsymbol{\epsilon}^{\mydprime},
\end{equation} 
where $\boldsymbol{\epsilon}^{\mydprime} \sim \mathcal{N}(0,\mathbf{I})$. Then, $\mathbf{z}$ will be decoded to compute the lattice lengths and angles, which then yield the lattice matrix ($\mathbf{L}_\mathbf{z}$), $N_a$, and $\mathbf{A}_\mathbf{z}$. In the original CDVAE, $\mathbf{A}_\mathbf{z}$ is the probability vector indicating the fraction of each atomic type in the compound and is used to perturb $Z$ by
\begin{equation}
    Z_t \sim \mathcal{M}(\text{softmax}(\mathbf{A} + \sigma_t^{\myprime}\mathbf{A}_\mathbf{z}))
\end{equation}
where $\mathcal{M}$ is a multinomial distribution, $\mathbf{A}$ is a one-hot vector of ground-truth atomic type $Z$, 
and $\sigma_t^{\myprime}$ is the variance for perturbing atomic types at time $t$, which is distinct from $\sigma_t$ used for perturbing the atomic coordinates. Similar to the original CDVAE, $\sigma_t^{\myprime}$ is selected from the range of [0.01, 5].

For the diffusion network, the input structures are constructed from $\mathbf{r}_{f_t}$, $Z_t$, and $\mathbf{L}_\mathbf{z}$ where the (Cartesian) atomic coordinates at time $t$ are computed by $\mathbf{r}_{c_t} = \mathbf{L}_\mathbf{z}\mathbf{r}_{f_t}$. These are then utilized by the crystal graphs for the diffusion network, whose node features are $\mathbf{f}_{t} = (Z_t, \mathbf{F}_t, \mathbf{z}, t)$ where $\mathbf{F}_t$ is a Fourier embedding feature of $\mathbf{r}_t$ (see section~\ref{sec:fourier}). As proposed by Ho et al. \cite{Ho2020}, we use the simple loss to train the model such that
\begin{equation}
    \mathcal{L}_{simple} = \|\boldsymbol{\epsilon} - \boldsymbol{\epsilon}_{\theta}(\mathbf{r}_{c_t}, \mathbf{f}_{t})\|^2.
\end{equation}
Since the diffusion model is trained to predict both $\boldsymbol{\epsilon}$ and $\mathbf{A}$, so the loss of the diffusion network is
\begin{equation}
    \mathcal{L}_{diff} = \mathcal{L}_{simple} + \lambda \mathcal{L}_{CE}(\mathbf{A},\mathbf{A}_{\theta}(\mathbf{r}_{c_t}, \mathbf{f}_{t})),
\end{equation}
where $\mathcal{L}_{CE}$ is the cross entropy loss, $\lambda$ is a loss scaling factor, and $t \in \{1, ..., T\}$ where $T=1000$. In this work, $t$ is randomly chosen for each crystal graph and randomly reinitialized for each epoch in the training process. The total loss in the trainig process is shown in Eq.~\ref{eq:total_loss}. 

In the reverse diffusion process, we measure the model performance of two tasks: reconstruction and generation tasks. For the former task, $\mathbf{z}$ is obtained from Eq.~\eqref{eq:reparam} by using the ground-truth structure as an input of the encoders. For the latter task, $\mathbf{z} \sim \mathcal{N}(0, \mathbf{I})$, which is then used to predict $N_a$, $\mathbf{L}_\mathbf{z}$, $\mathbf{A}_\mathbf{z}$, and concatenate with the node feature of the crystal graph in the diffusion network. At the initial step, $t=T$, $Z_T$ is sampled from the highest probability of $\mathbf{A}_\mathbf{z}$, and the final-time coordinate is obtained from sampling a Gaussian distribution, i.e. $\mathbf{r}_T \sim \mathcal{N}(0, \mathbf{I}).$ The coordinates can be denoised using Eq.~\eqref{eq:rf_t-1}, and the predicted atomic types are updated in each reversed time step by $\text{argmax}(\mathbf{A}_{\theta})$. 

\subsection{DFT calculations}
The Vienna \textit{ab initio} Simulation Package (VASP) was employed for structural relaxations and energy calculations based on DFT \cite{Kresse1996-1, Kresse1996-2}. The calculations were conducted under the generalized gradient approximation (GGA) and the project augmented wave (PAW) method \cite{Perdew1996, Blochl1994}. The thresholds for energy and force convergence were set to $10^{-5}$ eV and $10^{-5}$ eV/{\AA}, respectively. The plane-wave energy cutoff was set to 800 eV, and the Brillouin zone integration was carried out on a k-point mesh of 5 × 5 × 5 created by the Monkhorst-Pack method \cite{Monkhorst1976, Pack1977}.

\begin{acknowledgments}
This research project is supported by the Second Century Fund (C2F), Chulalongkorn University. This Research is funded by Thailand Science research and Innovation Fund Chulalongkorn University (IND66230002) and National Research Council of Thailand (NRCT): (NRCT5-RSA63001-04). T.C. acknowledges funding support from the NSRF via the Program Management Unit for Human Resources and Institutional Development, Research and Innovation [grant number B05F650024]. The authors acknowledge high performance computing resources including NVIDIA A100 GPUs from Chula Intelligent and Complex Systems Lab, Faculty of Science,
and from the Center for AI in Medicine (CU-AIM), Faculty of Medicine, Chulalongkorn University, Thailand. We acknowledge the supporting computing infrastructure provided by NSTDA, CU, CUAASC, NSRF via PMUB [B05F650021, B37G660013] (Thailand). URL:www.e-science.in.th. The Computational Materials Physics (CMP) Project, SLRI, Thailand, is acknowledged for providing computational resource.
\end{acknowledgments}

\nocite{*}

\bibliography{ref} 

\providecommand{\noopsort}[1]{}\providecommand{\singleletter}[1]{#1}%
\begin{thebibliography}{49}%
\makeatletter
\providecommand \@ifxundefined [1]{%
 \@ifx{#1\undefined}
}%
\providecommand \@ifnum [1]{%
 \ifnum #1\expandafter \@firstoftwo
 \else \expandafter \@secondoftwo
 \fi
}%
\providecommand \@ifx [1]{%
 \ifx #1\expandafter \@firstoftwo
 \else \expandafter \@secondoftwo
 \fi
}%
\providecommand \natexlab [1]{#1}%
\providecommand \enquote  [1]{``#1''}%
\providecommand \bibnamefont  [1]{#1}%
\providecommand \bibfnamefont [1]{#1}%
\providecommand \citenamefont [1]{#1}%
\providecommand \href@noop [0]{\@secondoftwo}%
\providecommand \href [0]{\begingroup \@sanitize@url \@href}%
\providecommand \@href[1]{\@@startlink{#1}\@@href}%
\providecommand \@@href[1]{\endgroup#1\@@endlink}%
\providecommand \@sanitize@url [0]{\catcode `\\12\catcode `\$12\catcode
  `\&12\catcode `\#12\catcode `\^12\catcode `\_12\catcode `\%12\relax}%
\providecommand \@@startlink[1]{}%
\providecommand \@@endlink[0]{}%
\providecommand \url  [0]{\begingroup\@sanitize@url \@url }%
\providecommand \@url [1]{\endgroup\@href {#1}{\urlprefix }}%
\providecommand \urlprefix  [0]{URL }%
\providecommand \Eprint [0]{\href }%
\providecommand \doibase [0]{https://doi.org/}%
\providecommand \selectlanguage [0]{\@gobble}%
\providecommand \bibinfo  [0]{\@secondoftwo}%
\providecommand \bibfield  [0]{\@secondoftwo}%
\providecommand \translation [1]{[#1]}%
\providecommand \BibitemOpen [0]{}%
\providecommand \bibitemStop [0]{}%
\providecommand \bibitemNoStop [0]{.\EOS\space}%
\providecommand \EOS [0]{\spacefactor3000\relax}%
\providecommand \BibitemShut  [1]{\csname bibitem#1\endcsname}%
\let\auto@bib@innerbib\@empty
\bibitem [{\citenamefont {Needs}\ and\ \citenamefont
  {Pickard}(2016)}]{Needs2016}%
  \BibitemOpen
  \bibfield  {author} {\bibinfo {author} {\bibfnamefont {R.~J.}\ \bibnamefont
  {Needs}}\ and\ \bibinfo {author} {\bibfnamefont {C.~J.}\ \bibnamefont
  {Pickard}},\ }\bibfield  {title} {\bibinfo {title} {{Perspective: Role of
  structure prediction in materials discovery and design}},\ }\href
  {https://doi.org/10.1063/1.4949361} {\bibfield  {journal} {\bibinfo
  {journal} {APL Materials}\ }\textbf {\bibinfo {volume} {4}},\ \bibinfo
  {pages} {053210} (\bibinfo {year} {2016})}\BibitemShut {NoStop}%
\bibitem [{\citenamefont {Kohn}\ and\ \citenamefont {Sham}(1965)}]{Kohn1965}%
  \BibitemOpen
  \bibfield  {author} {\bibinfo {author} {\bibfnamefont {W.}~\bibnamefont
  {Kohn}}\ and\ \bibinfo {author} {\bibfnamefont {L.~J.}\ \bibnamefont
  {Sham}},\ }\href@noop {} {\bibfield  {journal} {\bibinfo  {journal} {Phys.
  Rev.}\ }\textbf {\bibinfo {volume} {140}},\ \bibinfo {pages} {A1133}
  (\bibinfo {year} {1965})}\BibitemShut {NoStop}%
\bibitem [{\citenamefont {Oganov}\ and\ \citenamefont
  {Glass}(2006)}]{Uspex2006}%
  \BibitemOpen
  \bibfield  {author} {\bibinfo {author} {\bibfnamefont {A.~R.}\ \bibnamefont
  {Oganov}}\ and\ \bibinfo {author} {\bibfnamefont {C.~W.}\ \bibnamefont
  {Glass}},\ }\bibfield  {title} {\bibinfo {title} {{Crystal structure
  prediction using ab initio evolutionary techniques: Principles and
  applications}},\ }\href {https://doi.org/10.1063/1.2210932} {\bibfield
  {journal} {\bibinfo  {journal} {The Journal of Chemical Physics}\ }\textbf
  {\bibinfo {volume} {124}},\ \bibinfo {pages} {244704} (\bibinfo {year}
  {2006})}\BibitemShut {NoStop}%
\bibitem [{\citenamefont {Wang}\ \emph {et~al.}(2010)\citenamefont {Wang},
  \citenamefont {Lv}, \citenamefont {Zhu},\ and\ \citenamefont
  {Ma}}]{Calypso2010}%
  \BibitemOpen
  \bibfield  {author} {\bibinfo {author} {\bibfnamefont {Y.}~\bibnamefont
  {Wang}}, \bibinfo {author} {\bibfnamefont {J.}~\bibnamefont {Lv}}, \bibinfo
  {author} {\bibfnamefont {L.}~\bibnamefont {Zhu}},\ and\ \bibinfo {author}
  {\bibfnamefont {Y.}~\bibnamefont {Ma}},\ }\bibfield  {title} {\bibinfo
  {title} {Crystal structure prediction via particle-swarm optimization},\
  }\href {https://doi.org/10.1103/PhysRevB.82.094116} {\bibfield  {journal}
  {\bibinfo  {journal} {Phys. Rev. B}\ }\textbf {\bibinfo {volume} {82}},\
  \bibinfo {pages} {094116} (\bibinfo {year} {2010})}\BibitemShut {NoStop}%
\bibitem [{\citenamefont {Pickard}\ and\ \citenamefont
  {Needs}(2011)}]{Pickard2011}%
  \BibitemOpen
  \bibfield  {author} {\bibinfo {author} {\bibfnamefont {C.~J.}\ \bibnamefont
  {Pickard}}\ and\ \bibinfo {author} {\bibfnamefont {R.~J.}\ \bibnamefont
  {Needs}},\ }\bibfield  {title} {\bibinfo {title} {Ab initio random structure
  searching},\ }\href {https://doi.org/10.1088/0953-8984/23/5/053201}
  {\bibfield  {journal} {\bibinfo  {journal} {Journal of Physics: Condensed
  Matter}\ }\textbf {\bibinfo {volume} {23}},\ \bibinfo {pages} {053201}
  (\bibinfo {year} {2011})}\BibitemShut {NoStop}%
\bibitem [{\citenamefont {Oganov}\ \emph {et~al.}(2019)\citenamefont {Oganov},
  \citenamefont {Pickard}, \citenamefont {Zhu},\ and\ \citenamefont
  {Needs}}]{Oganov2019-dw}%
  \BibitemOpen
  \bibfield  {author} {\bibinfo {author} {\bibfnamefont {A.~R.}\ \bibnamefont
  {Oganov}}, \bibinfo {author} {\bibfnamefont {C.~J.}\ \bibnamefont {Pickard}},
  \bibinfo {author} {\bibfnamefont {Q.}~\bibnamefont {Zhu}},\ and\ \bibinfo
  {author} {\bibfnamefont {R.~J.}\ \bibnamefont {Needs}},\ }\bibfield  {title}
  {\bibinfo {title} {Structure prediction drives materials discovery},\
  }\href@noop {} {\bibfield  {journal} {\bibinfo  {journal} {Nature Reviews
  Materials}\ }\textbf {\bibinfo {volume} {4}},\ \bibinfo {pages} {331}
  (\bibinfo {year} {2019})}\BibitemShut {NoStop}%
\bibitem [{\citenamefont {Schön}\ \emph {et~al.}(2010)\citenamefont {Schön},
  \citenamefont {Doll},\ and\ \citenamefont {Jansen}}]{Schon2010}%
  \BibitemOpen
  \bibfield  {author} {\bibinfo {author} {\bibfnamefont {J.~C.}\ \bibnamefont
  {Schön}}, \bibinfo {author} {\bibfnamefont {K.}~\bibnamefont {Doll}},\ and\
  \bibinfo {author} {\bibfnamefont {M.}~\bibnamefont {Jansen}},\ }\bibfield
  {title} {\bibinfo {title} {Predicting solid compounds via global exploration
  of the energy landscape of solids on the ab initio level without recourse to
  experimental information},\ }\href
  {https://doi.org/https://doi.org/10.1002/pssb.200945246} {\bibfield
  {journal} {\bibinfo  {journal} {physica status solidi (b)}\ }\textbf
  {\bibinfo {volume} {247}},\ \bibinfo {pages} {23} (\bibinfo {year}
  {2010})}\BibitemShut {NoStop}%
\bibitem [{\citenamefont {Xie}\ \emph {et~al.}(2022)\citenamefont {Xie},
  \citenamefont {Fu}, \citenamefont {Ganea}, \citenamefont {Barzilay},\ and\
  \citenamefont {Jaakkola}}]{Xie2022cdvae}%
  \BibitemOpen
  \bibfield  {author} {\bibinfo {author} {\bibfnamefont {T.}~\bibnamefont
  {Xie}}, \bibinfo {author} {\bibfnamefont {X.}~\bibnamefont {Fu}}, \bibinfo
  {author} {\bibfnamefont {O.-E.}\ \bibnamefont {Ganea}}, \bibinfo {author}
  {\bibfnamefont {R.}~\bibnamefont {Barzilay}},\ and\ \bibinfo {author}
  {\bibfnamefont {T.~S.}\ \bibnamefont {Jaakkola}},\ }\bibfield  {title}
  {\bibinfo {title} {Crystal diffusion variational autoencoder for periodic
  material generation},\ }in\ \href
  {https://openreview.net/forum?id=03RLpj-tc_} {\emph {\bibinfo {booktitle}
  {International Conference on Learning Representations}}}\ (\bibinfo {year}
  {2022})\BibitemShut {NoStop}%
\bibitem [{\citenamefont {Shi}\ \emph {et~al.}(2021)\citenamefont {Shi},
  \citenamefont {Luo}, \citenamefont {Xu},\ and\ \citenamefont
  {Tang}}]{Shi2021}%
  \BibitemOpen
  \bibfield  {author} {\bibinfo {author} {\bibfnamefont {C.}~\bibnamefont
  {Shi}}, \bibinfo {author} {\bibfnamefont {S.}~\bibnamefont {Luo}}, \bibinfo
  {author} {\bibfnamefont {M.}~\bibnamefont {Xu}},\ and\ \bibinfo {author}
  {\bibfnamefont {J.}~\bibnamefont {Tang}},\ }\bibfield  {title} {\bibinfo
  {title} {Learning gradient fields for molecular conformation generation},\
  }in\ \href@noop {} {\emph {\bibinfo {booktitle} {Proceedings of the 38th
  International Conference on Machine Learning}}},\ \bibinfo {series}
  {Proceedings of Machine Learning Research}, Vol.\ \bibinfo {volume} {139},\
  \bibinfo {editor} {edited by\ \bibinfo {editor} {\bibfnamefont
  {M.}~\bibnamefont {Meila}}\ and\ \bibinfo {editor} {\bibfnamefont
  {T.}~\bibnamefont {Zhang}}}\ (\bibinfo  {publisher} {PMLR},\ \bibinfo {year}
  {2021})\ pp.\ \bibinfo {pages} {9558--9568}\BibitemShut {NoStop}%
\bibitem [{\citenamefont {Xu}\ \emph {et~al.}(2022)\citenamefont {Xu},
  \citenamefont {Yu}, \citenamefont {Song}, \citenamefont {Shi}, \citenamefont
  {Ermon},\ and\ \citenamefont {Tang}}]{Xu2022geodiff}%
  \BibitemOpen
  \bibfield  {author} {\bibinfo {author} {\bibfnamefont {M.}~\bibnamefont
  {Xu}}, \bibinfo {author} {\bibfnamefont {L.}~\bibnamefont {Yu}}, \bibinfo
  {author} {\bibfnamefont {Y.}~\bibnamefont {Song}}, \bibinfo {author}
  {\bibfnamefont {C.}~\bibnamefont {Shi}}, \bibinfo {author} {\bibfnamefont
  {S.}~\bibnamefont {Ermon}},\ and\ \bibinfo {author} {\bibfnamefont
  {J.}~\bibnamefont {Tang}},\ }\bibfield  {title} {\bibinfo {title} {Geodiff: A
  geometric diffusion model for molecular conformation generation},\ }in\ \href
  {https://openreview.net/forum?id=PzcvxEMzvQC} {\emph {\bibinfo {booktitle}
  {International Conference on Learning Representations}}}\ (\bibinfo {year}
  {2022})\BibitemShut {NoStop}%
\bibitem [{\citenamefont {Guan}\ \emph {et~al.}(2023)\citenamefont {Guan},
  \citenamefont {Qian}, \citenamefont {Peng}, \citenamefont {Su}, \citenamefont
  {Peng},\ and\ \citenamefont {Ma}}]{Guan2023edm}%
  \BibitemOpen
  \bibfield  {author} {\bibinfo {author} {\bibfnamefont {J.}~\bibnamefont
  {Guan}}, \bibinfo {author} {\bibfnamefont {W.~W.}\ \bibnamefont {Qian}},
  \bibinfo {author} {\bibfnamefont {X.}~\bibnamefont {Peng}}, \bibinfo {author}
  {\bibfnamefont {Y.}~\bibnamefont {Su}}, \bibinfo {author} {\bibfnamefont
  {J.}~\bibnamefont {Peng}},\ and\ \bibinfo {author} {\bibfnamefont
  {J.}~\bibnamefont {Ma}},\ }\bibfield  {title} {\bibinfo {title} {3d
  equivariant diffusion for target-aware molecule generation and affinity
  prediction},\ }in\ \href {https://openreview.net/forum?id=kJqXEPXMsE0} {\emph
  {\bibinfo {booktitle} {The Eleventh International Conference on Learning
  Representations}}}\ (\bibinfo {year} {2023})\BibitemShut {NoStop}%
\bibitem [{\citenamefont {Kang}\ and\ \citenamefont {Cho}(2019)}]{Kang2019}%
  \BibitemOpen
  \bibfield  {author} {\bibinfo {author} {\bibfnamefont {S.}~\bibnamefont
  {Kang}}\ and\ \bibinfo {author} {\bibfnamefont {K.}~\bibnamefont {Cho}},\
  }\bibfield  {title} {\bibinfo {title} {Conditional molecular design with deep
  generative models},\ }\href {https://doi.org/10.1021/acs.jcim.8b00263}
  {\bibfield  {journal} {\bibinfo  {journal} {Journal of Chemical Information
  and Modeling}\ }\textbf {\bibinfo {volume} {59}},\ \bibinfo {pages} {43}
  (\bibinfo {year} {2019})},\ \bibinfo {note} {pMID: 30016587}\BibitemShut
  {NoStop}%
\bibitem [{\citenamefont {Lim}\ \emph {et~al.}(2018)\citenamefont {Lim},
  \citenamefont {Ryu}, \citenamefont {Kim},\ and\ \citenamefont
  {Kim}}]{Lim2018-mt}%
  \BibitemOpen
  \bibfield  {author} {\bibinfo {author} {\bibfnamefont {J.}~\bibnamefont
  {Lim}}, \bibinfo {author} {\bibfnamefont {S.}~\bibnamefont {Ryu}}, \bibinfo
  {author} {\bibfnamefont {J.~W.}\ \bibnamefont {Kim}},\ and\ \bibinfo {author}
  {\bibfnamefont {W.~Y.}\ \bibnamefont {Kim}},\ }\bibfield  {title} {\bibinfo
  {title} {Molecular generative model based on conditional variational
  autoencoder for de novo molecular design},\ }\href@noop {} {\bibfield
  {journal} {\bibinfo  {journal} {Journal of Cheminformatics}\ }\textbf
  {\bibinfo {volume} {10}},\ \bibinfo {pages} {31} (\bibinfo {year}
  {2018})}\BibitemShut {NoStop}%
\bibitem [{\citenamefont {Song}\ \emph {et~al.}(2022)\citenamefont {Song},
  \citenamefont {Shen}, \citenamefont {Xing},\ and\ \citenamefont
  {Ermon}}]{Song2022solving}%
  \BibitemOpen
  \bibfield  {author} {\bibinfo {author} {\bibfnamefont {Y.}~\bibnamefont
  {Song}}, \bibinfo {author} {\bibfnamefont {L.}~\bibnamefont {Shen}}, \bibinfo
  {author} {\bibfnamefont {L.}~\bibnamefont {Xing}},\ and\ \bibinfo {author}
  {\bibfnamefont {S.}~\bibnamefont {Ermon}},\ }\bibfield  {title} {\bibinfo
  {title} {Solving inverse problems in medical imaging with score-based
  generative models},\ }in\ \href {https://openreview.net/forum?id=vaRCHVj0uGI}
  {\emph {\bibinfo {booktitle} {International Conference on Learning
  Representations}}}\ (\bibinfo {year} {2022})\BibitemShut {NoStop}%
\bibitem [{\citenamefont {Cui}\ \emph {et~al.}(2019)\citenamefont {Cui},
  \citenamefont {Jiang}, \citenamefont {Jiang}, \citenamefont {Shang},
  \citenamefont {Zhu}, \citenamefont {Hu}, \citenamefont {Xu},\ and\
  \citenamefont {Chu}}]{Cui2019}%
  \BibitemOpen
  \bibfield  {author} {\bibinfo {author} {\bibfnamefont {A.}~\bibnamefont
  {Cui}}, \bibinfo {author} {\bibfnamefont {K.}~\bibnamefont {Jiang}}, \bibinfo
  {author} {\bibfnamefont {M.}~\bibnamefont {Jiang}}, \bibinfo {author}
  {\bibfnamefont {L.}~\bibnamefont {Shang}}, \bibinfo {author} {\bibfnamefont
  {L.}~\bibnamefont {Zhu}}, \bibinfo {author} {\bibfnamefont {Z.}~\bibnamefont
  {Hu}}, \bibinfo {author} {\bibfnamefont {G.}~\bibnamefont {Xu}},\ and\
  \bibinfo {author} {\bibfnamefont {J.}~\bibnamefont {Chu}},\ }\bibfield
  {title} {\bibinfo {title} {Decoding phases of matter by machine-learning
  raman spectroscopy},\ }\href
  {https://doi.org/10.1103/PhysRevApplied.12.054049} {\bibfield  {journal}
  {\bibinfo  {journal} {Phys. Rev. Appl.}\ }\textbf {\bibinfo {volume} {12}},\
  \bibinfo {pages} {054049} (\bibinfo {year} {2019})}\BibitemShut {NoStop}%
\bibitem [{\citenamefont {Carbone}\ \emph {et~al.}(2020)\citenamefont
  {Carbone}, \citenamefont {Topsakal}, \citenamefont {Lu},\ and\ \citenamefont
  {Yoo}}]{Carbone2020}%
  \BibitemOpen
  \bibfield  {author} {\bibinfo {author} {\bibfnamefont {M.~R.}\ \bibnamefont
  {Carbone}}, \bibinfo {author} {\bibfnamefont {M.}~\bibnamefont {Topsakal}},
  \bibinfo {author} {\bibfnamefont {D.}~\bibnamefont {Lu}},\ and\ \bibinfo
  {author} {\bibfnamefont {S.}~\bibnamefont {Yoo}},\ }\bibfield  {title}
  {\bibinfo {title} {Machine-learning x-ray absorption spectra to quantitative
  accuracy},\ }\href {https://doi.org/10.1103/PhysRevLett.124.156401}
  {\bibfield  {journal} {\bibinfo  {journal} {Phys. Rev. Lett.}\ }\textbf
  {\bibinfo {volume} {124}},\ \bibinfo {pages} {156401} (\bibinfo {year}
  {2020})}\BibitemShut {NoStop}%
\bibitem [{\citenamefont {Liang}\ \emph {et~al.}(2023)\citenamefont {Liang},
  \citenamefont {Carbone}, \citenamefont {Chen}, \citenamefont {Meng},
  \citenamefont {Stavitski}, \citenamefont {Lu}, \citenamefont {Hybertsen},\
  and\ \citenamefont {Qu}}]{Liang2023}%
  \BibitemOpen
  \bibfield  {author} {\bibinfo {author} {\bibfnamefont {Z.}~\bibnamefont
  {Liang}}, \bibinfo {author} {\bibfnamefont {M.~R.}\ \bibnamefont {Carbone}},
  \bibinfo {author} {\bibfnamefont {W.}~\bibnamefont {Chen}}, \bibinfo {author}
  {\bibfnamefont {F.}~\bibnamefont {Meng}}, \bibinfo {author} {\bibfnamefont
  {E.}~\bibnamefont {Stavitski}}, \bibinfo {author} {\bibfnamefont
  {D.}~\bibnamefont {Lu}}, \bibinfo {author} {\bibfnamefont {M.~S.}\
  \bibnamefont {Hybertsen}},\ and\ \bibinfo {author} {\bibfnamefont
  {X.}~\bibnamefont {Qu}},\ }\bibfield  {title} {\bibinfo {title} {Decoding
  structure-spectrum relationships with physically organized latent spaces},\
  }\href {https://doi.org/10.1103/PhysRevMaterials.7.053802} {\bibfield
  {journal} {\bibinfo  {journal} {Phys. Rev. Mater.}\ }\textbf {\bibinfo
  {volume} {7}},\ \bibinfo {pages} {053802} (\bibinfo {year}
  {2023})}\BibitemShut {NoStop}%
\bibitem [{\citenamefont {Song}\ and\ \citenamefont {Ermon}(2019)}]{Song2019}%
  \BibitemOpen
  \bibfield  {author} {\bibinfo {author} {\bibfnamefont {Y.}~\bibnamefont
  {Song}}\ and\ \bibinfo {author} {\bibfnamefont {S.}~\bibnamefont {Ermon}},\
  }\bibfield  {title} {\bibinfo {title} {Generative modeling by estimating
  gradients of the data distribution},\ }in\ \href
  {https://proceedings.neurips.cc/paper_files/paper/2019/file/3001ef257407d5a371a96dcd947c7d93-Paper.pdf}
  {\emph {\bibinfo {booktitle} {Advances in Neural Information Processing
  Systems}}},\ Vol.~\bibinfo {volume} {32},\ \bibinfo {editor} {edited by\
  \bibinfo {editor} {\bibfnamefont {H.}~\bibnamefont {Wallach}}, \bibinfo
  {editor} {\bibfnamefont {H.}~\bibnamefont {Larochelle}}, \bibinfo {editor}
  {\bibfnamefont {A.}~\bibnamefont {Beygelzimer}}, \bibinfo {editor}
  {\bibfnamefont {F.}~\bibnamefont {d\textquotesingle Alch\'{e}-Buc}}, \bibinfo
  {editor} {\bibfnamefont {E.}~\bibnamefont {Fox}},\ and\ \bibinfo {editor}
  {\bibfnamefont {R.}~\bibnamefont {Garnett}}}\ (\bibinfo  {publisher} {Curran
  Associates, Inc.},\ \bibinfo {year} {2019})\BibitemShut {NoStop}%
\bibitem [{\citenamefont {Song}\ \emph {et~al.}(2021)\citenamefont {Song},
  \citenamefont {Sohl-Dickstein}, \citenamefont {Kingma}, \citenamefont
  {Kumar}, \citenamefont {Ermon},\ and\ \citenamefont {Poole}}]{Song2021}%
  \BibitemOpen
  \bibfield  {author} {\bibinfo {author} {\bibfnamefont {Y.}~\bibnamefont
  {Song}}, \bibinfo {author} {\bibfnamefont {J.}~\bibnamefont
  {Sohl-Dickstein}}, \bibinfo {author} {\bibfnamefont {D.~P.}\ \bibnamefont
  {Kingma}}, \bibinfo {author} {\bibfnamefont {A.}~\bibnamefont {Kumar}},
  \bibinfo {author} {\bibfnamefont {S.}~\bibnamefont {Ermon}},\ and\ \bibinfo
  {author} {\bibfnamefont {B.}~\bibnamefont {Poole}},\ }\bibfield  {title}
  {\bibinfo {title} {Score-based generative modeling through stochastic
  differential equations},\ }in\ \href
  {https://openreview.net/forum?id=PxTIG12RRHS} {\emph {\bibinfo {booktitle}
  {International Conference on Learning Representations}}}\ (\bibinfo {year}
  {2021})\BibitemShut {NoStop}%
\bibitem [{\citenamefont {Ho}\ \emph {et~al.}(2020)\citenamefont {Ho},
  \citenamefont {Jain},\ and\ \citenamefont {Abbeel}}]{Ho2020}%
  \BibitemOpen
  \bibfield  {author} {\bibinfo {author} {\bibfnamefont {J.}~\bibnamefont
  {Ho}}, \bibinfo {author} {\bibfnamefont {A.}~\bibnamefont {Jain}},\ and\
  \bibinfo {author} {\bibfnamefont {P.}~\bibnamefont {Abbeel}},\ }\bibfield
  {title} {\bibinfo {title} {Denoising diffusion probabilistic models},\ }in\
  \href
  {https://proceedings.neurips.cc/paper_files/paper/2020/file/4c5bcfec8584af0d967f1ab10179ca4b-Paper.pdf}
  {\emph {\bibinfo {booktitle} {Advances in Neural Information Processing
  Systems}}},\ Vol.~\bibinfo {volume} {33},\ \bibinfo {editor} {edited by\
  \bibinfo {editor} {\bibfnamefont {H.}~\bibnamefont {Larochelle}}, \bibinfo
  {editor} {\bibfnamefont {M.}~\bibnamefont {Ranzato}}, \bibinfo {editor}
  {\bibfnamefont {R.}~\bibnamefont {Hadsell}}, \bibinfo {editor} {\bibfnamefont
  {M.}~\bibnamefont {Balcan}},\ and\ \bibinfo {editor} {\bibfnamefont
  {H.}~\bibnamefont {Lin}}}\ (\bibinfo  {publisher} {Curran Associates, Inc.},\
  \bibinfo {year} {2020})\ pp.\ \bibinfo {pages} {6840--6851}\BibitemShut
  {NoStop}%
\bibitem [{\citenamefont {Bronstein}\ \emph {et~al.}(2021)\citenamefont
  {Bronstein}, \citenamefont {Bruna}, \citenamefont {Cohen},\ and\
  \citenamefont {Velickovic}}]{Bronstein2021}%
  \BibitemOpen
  \bibfield  {author} {\bibinfo {author} {\bibfnamefont {M.~M.}\ \bibnamefont
  {Bronstein}}, \bibinfo {author} {\bibfnamefont {J.}~\bibnamefont {Bruna}},
  \bibinfo {author} {\bibfnamefont {T.}~\bibnamefont {Cohen}},\ and\ \bibinfo
  {author} {\bibfnamefont {P.}~\bibnamefont {Velickovic}},\ }\bibfield  {title}
  {\bibinfo {title} {Geometric deep learning: Grids, groups, graphs, geodesics,
  and gauges},\ }\href {https://arxiv.org/abs/2104.13478} {\bibfield  {journal}
  {\bibinfo  {journal} {CoRR}\ }\textbf {\bibinfo {volume} {abs/2104.13478}}
  (\bibinfo {year} {2021})},\ \Eprint {https://arxiv.org/abs/2104.13478}
  {2104.13478} \BibitemShut {NoStop}%
\bibitem [{\citenamefont {Cohen}\ \emph {et~al.}(2018)\citenamefont {Cohen},
  \citenamefont {Geiger}, \citenamefont {Köhler},\ and\ \citenamefont
  {Welling}}]{Cohen2018}%
  \BibitemOpen
  \bibfield  {author} {\bibinfo {author} {\bibfnamefont {T.~S.}\ \bibnamefont
  {Cohen}}, \bibinfo {author} {\bibfnamefont {M.}~\bibnamefont {Geiger}},
  \bibinfo {author} {\bibfnamefont {J.}~\bibnamefont {Köhler}},\ and\ \bibinfo
  {author} {\bibfnamefont {M.}~\bibnamefont {Welling}},\ }\bibfield  {title}
  {\bibinfo {title} {Spherical {CNN}s},\ }in\ \href
  {https://openreview.net/forum?id=Hkbd5xZRb} {\emph {\bibinfo {booktitle}
  {International Conference on Learning Representations}}}\ (\bibinfo {year}
  {2018})\BibitemShut {NoStop}%
\bibitem [{\citenamefont {Thomas}\ \emph {et~al.}(2018)\citenamefont {Thomas},
  \citenamefont {Smidt}, \citenamefont {Kearnes}, \citenamefont {Yang},
  \citenamefont {Li}, \citenamefont {Kohlhoff},\ and\ \citenamefont
  {Riley}}]{Thomas2018}%
  \BibitemOpen
  \bibfield  {author} {\bibinfo {author} {\bibfnamefont {N.}~\bibnamefont
  {Thomas}}, \bibinfo {author} {\bibfnamefont {T.}~\bibnamefont {Smidt}},
  \bibinfo {author} {\bibfnamefont {S.}~\bibnamefont {Kearnes}}, \bibinfo
  {author} {\bibfnamefont {L.}~\bibnamefont {Yang}}, \bibinfo {author}
  {\bibfnamefont {L.}~\bibnamefont {Li}}, \bibinfo {author} {\bibfnamefont
  {K.}~\bibnamefont {Kohlhoff}},\ and\ \bibinfo {author} {\bibfnamefont
  {P.}~\bibnamefont {Riley}},\ }\href
  {https://doi.org/10.48550/ARXIV.1802.08219} {\bibinfo {title} {Tensor field
  networks: Rotation- and translation-equivariant neural networks for 3d point
  clouds}} (\bibinfo {year} {2018})\BibitemShut {NoStop}%
\bibitem [{\citenamefont {Kingma}\ and\ \citenamefont
  {Welling}(2014)}]{Kingma2014}%
  \BibitemOpen
  \bibfield  {author} {\bibinfo {author} {\bibfnamefont {D.~P.}\ \bibnamefont
  {Kingma}}\ and\ \bibinfo {author} {\bibfnamefont {M.}~\bibnamefont
  {Welling}},\ }\bibfield  {title} {\bibinfo {title} {Auto-encoding variational
  bayes},\ }in\ \href@noop {} {\emph {\bibinfo {booktitle} {International
  Conference on Learning Representations}}}\ (\bibinfo {year}
  {2014})\BibitemShut {NoStop}%
\bibitem [{\citenamefont {Jiao}\ \emph {et~al.}(2023)\citenamefont {Jiao},
  \citenamefont {Huang}, \citenamefont {Lin}, \citenamefont {Han},
  \citenamefont {Chen}, \citenamefont {Lu},\ and\ \citenamefont
  {Liu}}]{Jiao2023diffcsp}%
  \BibitemOpen
  \bibfield  {author} {\bibinfo {author} {\bibfnamefont {R.}~\bibnamefont
  {Jiao}}, \bibinfo {author} {\bibfnamefont {W.}~\bibnamefont {Huang}},
  \bibinfo {author} {\bibfnamefont {P.}~\bibnamefont {Lin}}, \bibinfo {author}
  {\bibfnamefont {J.}~\bibnamefont {Han}}, \bibinfo {author} {\bibfnamefont
  {P.}~\bibnamefont {Chen}}, \bibinfo {author} {\bibfnamefont {Y.}~\bibnamefont
  {Lu}},\ and\ \bibinfo {author} {\bibfnamefont {Y.}~\bibnamefont {Liu}},\
  }\bibfield  {title} {\bibinfo {title} {Crystal structure prediction by joint
  equivariant diffusion on lattices and fractional coordinates},\ }in\ \href
  {https://openreview.net/forum?id=VPByphdu24j} {\emph {\bibinfo {booktitle}
  {Workshop on ''Machine Learning for Materials'' ICLR 2023}}}\ (\bibinfo
  {year} {2023})\BibitemShut {NoStop}%
\bibitem [{\citenamefont {Okhotin}\ \emph {et~al.}(2023)\citenamefont
  {Okhotin}, \citenamefont {Molchanov}, \citenamefont {Arkhipkin},
  \citenamefont {Bartosh}, \citenamefont {Alanov},\ and\ \citenamefont
  {Vetrov}}]{Okhotin2023starshaped}%
  \BibitemOpen
  \bibfield  {author} {\bibinfo {author} {\bibfnamefont {A.}~\bibnamefont
  {Okhotin}}, \bibinfo {author} {\bibfnamefont {D.}~\bibnamefont {Molchanov}},
  \bibinfo {author} {\bibfnamefont {V.}~\bibnamefont {Arkhipkin}}, \bibinfo
  {author} {\bibfnamefont {G.}~\bibnamefont {Bartosh}}, \bibinfo {author}
  {\bibfnamefont {A.}~\bibnamefont {Alanov}},\ and\ \bibinfo {author}
  {\bibfnamefont {D.}~\bibnamefont {Vetrov}},\ }\href@noop {} {\bibinfo {title}
  {Star-shaped denoising diffusion probabilistic models}} (\bibinfo {year}
  {2023}),\ \Eprint {https://arxiv.org/abs/2302.05259} {arXiv:2302.05259
  [stat.ML]} \BibitemShut {NoStop}%
\bibitem [{\citenamefont {Gasteiger}\ \emph {et~al.}(2022)\citenamefont
  {Gasteiger}, \citenamefont {Giri}, \citenamefont {Margraf},\ and\
  \citenamefont {Günnemann}}]{Gasteiger2022dimenetpp}%
  \BibitemOpen
  \bibfield  {author} {\bibinfo {author} {\bibfnamefont {J.}~\bibnamefont
  {Gasteiger}}, \bibinfo {author} {\bibfnamefont {S.}~\bibnamefont {Giri}},
  \bibinfo {author} {\bibfnamefont {J.~T.}\ \bibnamefont {Margraf}},\ and\
  \bibinfo {author} {\bibfnamefont {S.}~\bibnamefont {Günnemann}},\
  }\href@noop {} {\bibinfo {title} {Fast and uncertainty-aware directional
  message passing for non-equilibrium molecules}} (\bibinfo {year} {2022}),\
  \Eprint {https://arxiv.org/abs/2011.14115} {arXiv:2011.14115 [cs.LG]}
  \BibitemShut {NoStop}%
\bibitem [{\citenamefont {Hu*}\ \emph {et~al.}(2020)\citenamefont {Hu*},
  \citenamefont {Liu*}, \citenamefont {Gomes}, \citenamefont {Zitnik},
  \citenamefont {Liang}, \citenamefont {Pande},\ and\ \citenamefont
  {Leskovec}}]{Hu2020gine}%
  \BibitemOpen
  \bibfield  {author} {\bibinfo {author} {\bibfnamefont {W.}~\bibnamefont
  {Hu*}}, \bibinfo {author} {\bibfnamefont {B.}~\bibnamefont {Liu*}}, \bibinfo
  {author} {\bibfnamefont {J.}~\bibnamefont {Gomes}}, \bibinfo {author}
  {\bibfnamefont {M.}~\bibnamefont {Zitnik}}, \bibinfo {author} {\bibfnamefont
  {P.}~\bibnamefont {Liang}}, \bibinfo {author} {\bibfnamefont
  {V.}~\bibnamefont {Pande}},\ and\ \bibinfo {author} {\bibfnamefont
  {J.}~\bibnamefont {Leskovec}},\ }\bibfield  {title} {\bibinfo {title}
  {Strategies for pre-training graph neural networks},\ }in\ \href
  {https://openreview.net/forum?id=HJlWWJSFDH} {\emph {\bibinfo {booktitle}
  {International Conference on Learning Representations}}}\ (\bibinfo {year}
  {2020})\BibitemShut {NoStop}%
\bibitem [{\citenamefont {Sch\"{u}tt}\ \emph {et~al.}(2017)\citenamefont
  {Sch\"{u}tt}, \citenamefont {Kindermans}, \citenamefont {Sauceda~Felix},
  \citenamefont {Chmiela}, \citenamefont {Tkatchenko},\ and\ \citenamefont
  {M\"{u}ller}}]{Schutt2017schnet}%
  \BibitemOpen
  \bibfield  {author} {\bibinfo {author} {\bibfnamefont {K.}~\bibnamefont
  {Sch\"{u}tt}}, \bibinfo {author} {\bibfnamefont {P.-J.}\ \bibnamefont
  {Kindermans}}, \bibinfo {author} {\bibfnamefont {H.~E.}\ \bibnamefont
  {Sauceda~Felix}}, \bibinfo {author} {\bibfnamefont {S.}~\bibnamefont
  {Chmiela}}, \bibinfo {author} {\bibfnamefont {A.}~\bibnamefont
  {Tkatchenko}},\ and\ \bibinfo {author} {\bibfnamefont {K.-R.}\ \bibnamefont
  {M\"{u}ller}},\ }\bibfield  {title} {\bibinfo {title} {Schnet: A
  continuous-filter convolutional neural network for modeling quantum
  interactions},\ }in\ \href
  {https://proceedings.neurips.cc/paper_files/paper/2017/file/303ed4c69846ab36c2904d3ba8573050-Paper.pdf}
  {\emph {\bibinfo {booktitle} {Advances in Neural Information Processing
  Systems}}},\ Vol.~\bibinfo {volume} {30},\ \bibinfo {editor} {edited by\
  \bibinfo {editor} {\bibfnamefont {I.}~\bibnamefont {Guyon}}, \bibinfo
  {editor} {\bibfnamefont {U.~V.}\ \bibnamefont {Luxburg}}, \bibinfo {editor}
  {\bibfnamefont {S.}~\bibnamefont {Bengio}}, \bibinfo {editor} {\bibfnamefont
  {H.}~\bibnamefont {Wallach}}, \bibinfo {editor} {\bibfnamefont
  {R.}~\bibnamefont {Fergus}}, \bibinfo {editor} {\bibfnamefont
  {S.}~\bibnamefont {Vishwanathan}},\ and\ \bibinfo {editor} {\bibfnamefont
  {R.}~\bibnamefont {Garnett}}}\ (\bibinfo  {publisher} {Curran Associates,
  Inc.},\ \bibinfo {year} {2017})\BibitemShut {NoStop}%
\bibitem [{\citenamefont {Castelli}\ \emph
  {et~al.}(2012{\natexlab{a}})\citenamefont {Castelli}, \citenamefont {Landis},
  \citenamefont {Thygesen}, \citenamefont {Dahl}, \citenamefont {Chorkendorff},
  \citenamefont {Jaramillo},\ and\ \citenamefont
  {Jacobsen}}]{Castelli2012perov5}%
  \BibitemOpen
  \bibfield  {author} {\bibinfo {author} {\bibfnamefont {I.~E.}\ \bibnamefont
  {Castelli}}, \bibinfo {author} {\bibfnamefont {D.~D.}\ \bibnamefont
  {Landis}}, \bibinfo {author} {\bibfnamefont {K.~S.}\ \bibnamefont
  {Thygesen}}, \bibinfo {author} {\bibfnamefont {S.}~\bibnamefont {Dahl}},
  \bibinfo {author} {\bibfnamefont {I.}~\bibnamefont {Chorkendorff}}, \bibinfo
  {author} {\bibfnamefont {T.~F.}\ \bibnamefont {Jaramillo}},\ and\ \bibinfo
  {author} {\bibfnamefont {K.~W.}\ \bibnamefont {Jacobsen}},\ }\bibfield
  {title} {\bibinfo {title} {New cubic perovskites for one-and two-photon water
  splitting using the computational materials repository},\ }\href@noop {}
  {\bibfield  {journal} {\bibinfo  {journal} {Energy \& Environmental Science}\
  }\textbf {\bibinfo {volume} {5}},\ \bibinfo {pages} {9034} (\bibinfo {year}
  {2012}{\natexlab{a}})}\BibitemShut {NoStop}%
\bibitem [{\citenamefont {Castelli}\ \emph
  {et~al.}(2012{\natexlab{b}})\citenamefont {Castelli}, \citenamefont {Olsen},
  \citenamefont {Datta}, \citenamefont {Landis}, \citenamefont {Dahl},
  \citenamefont {Thygesen},\ and\ \citenamefont
  {Jacobsen}}]{Castelli2012perov5-2}%
  \BibitemOpen
  \bibfield  {author} {\bibinfo {author} {\bibfnamefont {I.~E.}\ \bibnamefont
  {Castelli}}, \bibinfo {author} {\bibfnamefont {T.}~\bibnamefont {Olsen}},
  \bibinfo {author} {\bibfnamefont {S.}~\bibnamefont {Datta}}, \bibinfo
  {author} {\bibfnamefont {D.~D.}\ \bibnamefont {Landis}}, \bibinfo {author}
  {\bibfnamefont {S.}~\bibnamefont {Dahl}}, \bibinfo {author} {\bibfnamefont
  {K.~S.}\ \bibnamefont {Thygesen}},\ and\ \bibinfo {author} {\bibfnamefont
  {K.~W.}\ \bibnamefont {Jacobsen}},\ }\bibfield  {title} {\bibinfo {title}
  {Computational screening of perovskite metal oxides for optimal solar light
  capture},\ }\href@noop {} {\bibfield  {journal} {\bibinfo  {journal} {Energy
  \& Environmental Science}\ }\textbf {\bibinfo {volume} {5}},\ \bibinfo
  {pages} {5814} (\bibinfo {year} {2012}{\natexlab{b}})}\BibitemShut {NoStop}%
\bibitem [{\citenamefont {Pickard}()}]{Carbon2020data}%
  \BibitemOpen
  \bibfield  {author} {\bibinfo {author} {\bibfnamefont {C.~J.}\ \bibnamefont
  {Pickard}},\ }\href {https://doi.org/10.24435/MATERIALSCLOUD:2020.0026/V1}
  {}\BibitemShut {NoStop}%
\bibitem [{\citenamefont {Jain}\ \emph {et~al.}(2013)\citenamefont {Jain},
  \citenamefont {Ong}, \citenamefont {Hautier}, \citenamefont {Chen},
  \citenamefont {Richards}, \citenamefont {Dacek}, \citenamefont {Cholia},
  \citenamefont {Gunter}, \citenamefont {Skinner}, \citenamefont {Ceder} \emph
  {et~al.}}]{Jain2013mp20}%
  \BibitemOpen
  \bibfield  {author} {\bibinfo {author} {\bibfnamefont {A.}~\bibnamefont
  {Jain}}, \bibinfo {author} {\bibfnamefont {S.~P.}\ \bibnamefont {Ong}},
  \bibinfo {author} {\bibfnamefont {G.}~\bibnamefont {Hautier}}, \bibinfo
  {author} {\bibfnamefont {W.}~\bibnamefont {Chen}}, \bibinfo {author}
  {\bibfnamefont {W.~D.}\ \bibnamefont {Richards}}, \bibinfo {author}
  {\bibfnamefont {S.}~\bibnamefont {Dacek}}, \bibinfo {author} {\bibfnamefont
  {S.}~\bibnamefont {Cholia}}, \bibinfo {author} {\bibfnamefont
  {D.}~\bibnamefont {Gunter}}, \bibinfo {author} {\bibfnamefont
  {D.}~\bibnamefont {Skinner}}, \bibinfo {author} {\bibfnamefont
  {G.}~\bibnamefont {Ceder}}, \emph {et~al.},\ }\bibfield  {title} {\bibinfo
  {title} {Commentary: The materials project: A materials genome approach to
  accelerating materials innovation},\ }\href@noop {} {\bibfield  {journal}
  {\bibinfo  {journal} {APL materials}\ }\textbf {\bibinfo {volume} {1}},\
  \bibinfo {pages} {011002} (\bibinfo {year} {2013})}\BibitemShut {NoStop}%
\bibitem [{\citenamefont {Grosse-Kunstleve}\ \emph {et~al.}(2004)\citenamefont
  {Grosse-Kunstleve}, \citenamefont {Sauter},\ and\ \citenamefont
  {Adams}}]{Grosse-Kunstleve2004}%
  \BibitemOpen
  \bibfield  {author} {\bibinfo {author} {\bibfnamefont {R.~W.}\ \bibnamefont
  {Grosse-Kunstleve}}, \bibinfo {author} {\bibfnamefont {N.~K.}\ \bibnamefont
  {Sauter}},\ and\ \bibinfo {author} {\bibfnamefont {P.~D.}\ \bibnamefont
  {Adams}},\ }\bibfield  {title} {\bibinfo {title} {{Numerically stable
  algorithms for the computation of reduced unit cells}},\ }\href
  {https://doi.org/10.1107/S010876730302186X} {\bibfield  {journal} {\bibinfo
  {journal} {Acta Crystallographica Section A}\ }\textbf {\bibinfo {volume}
  {60}},\ \bibinfo {pages} {1} (\bibinfo {year} {2004})}\BibitemShut {NoStop}%
\bibitem [{\citenamefont {Ren}\ \emph {et~al.}(2022)\citenamefont {Ren},
  \citenamefont {Tian}, \citenamefont {Noh}, \citenamefont {Oviedo},
  \citenamefont {Xing}, \citenamefont {Li}, \citenamefont {Liang},
  \citenamefont {Zhu}, \citenamefont {Aberle}, \citenamefont {Sun},
  \citenamefont {Wang}, \citenamefont {Liu}, \citenamefont {Li}, \citenamefont
  {Jayavelu}, \citenamefont {Hippalgaonkar}, \citenamefont {Jung},\ and\
  \citenamefont {Buonassisi}}]{Ren2022}%
  \BibitemOpen
  \bibfield  {author} {\bibinfo {author} {\bibfnamefont {Z.}~\bibnamefont
  {Ren}}, \bibinfo {author} {\bibfnamefont {S.~I.~P.}\ \bibnamefont {Tian}},
  \bibinfo {author} {\bibfnamefont {J.}~\bibnamefont {Noh}}, \bibinfo {author}
  {\bibfnamefont {F.}~\bibnamefont {Oviedo}}, \bibinfo {author} {\bibfnamefont
  {G.}~\bibnamefont {Xing}}, \bibinfo {author} {\bibfnamefont {J.}~\bibnamefont
  {Li}}, \bibinfo {author} {\bibfnamefont {Q.}~\bibnamefont {Liang}}, \bibinfo
  {author} {\bibfnamefont {R.}~\bibnamefont {Zhu}}, \bibinfo {author}
  {\bibfnamefont {A.~G.}\ \bibnamefont {Aberle}}, \bibinfo {author}
  {\bibfnamefont {S.}~\bibnamefont {Sun}}, \bibinfo {author} {\bibfnamefont
  {X.}~\bibnamefont {Wang}}, \bibinfo {author} {\bibfnamefont {Y.}~\bibnamefont
  {Liu}}, \bibinfo {author} {\bibfnamefont {Q.}~\bibnamefont {Li}}, \bibinfo
  {author} {\bibfnamefont {S.}~\bibnamefont {Jayavelu}}, \bibinfo {author}
  {\bibfnamefont {K.}~\bibnamefont {Hippalgaonkar}}, \bibinfo {author}
  {\bibfnamefont {Y.}~\bibnamefont {Jung}},\ and\ \bibinfo {author}
  {\bibfnamefont {T.}~\bibnamefont {Buonassisi}},\ }\bibfield  {title}
  {\bibinfo {title} {An invertible crystallographic representation for general
  inverse design of inorganic crystals with targeted properties},\ }\href
  {https://doi.org/https://doi.org/10.1016/j.matt.2021.11.032} {\bibfield
  {journal} {\bibinfo  {journal} {Matter}\ }\textbf {\bibinfo {volume} {5}},\
  \bibinfo {pages} {314} (\bibinfo {year} {2022})}\BibitemShut {NoStop}%
\bibitem [{\citenamefont {Xu}\ \emph {et~al.}(2019)\citenamefont {Xu},
  \citenamefont {Hu}, \citenamefont {Leskovec},\ and\ \citenamefont
  {Jegelka}}]{Xu2018gin}%
  \BibitemOpen
  \bibfield  {author} {\bibinfo {author} {\bibfnamefont {K.}~\bibnamefont
  {Xu}}, \bibinfo {author} {\bibfnamefont {W.}~\bibnamefont {Hu}}, \bibinfo
  {author} {\bibfnamefont {J.}~\bibnamefont {Leskovec}},\ and\ \bibinfo
  {author} {\bibfnamefont {S.}~\bibnamefont {Jegelka}},\ }\bibfield  {title}
  {\bibinfo {title} {How powerful are graph neural networks?},\ }in\ \href
  {https://openreview.net/forum?id=ryGs6iA5Km} {\emph {\bibinfo {booktitle}
  {International Conference on Learning Representations}}}\ (\bibinfo {year}
  {2019})\BibitemShut {NoStop}%
\bibitem [{\citenamefont {Wu}\ \emph {et~al.}(2013)\citenamefont {Wu},
  \citenamefont {Lazic}, \citenamefont {Hautier}, \citenamefont {Persson},\
  and\ \citenamefont {Ceder}}]{Wu2013}%
  \BibitemOpen
  \bibfield  {author} {\bibinfo {author} {\bibfnamefont {Y.}~\bibnamefont
  {Wu}}, \bibinfo {author} {\bibfnamefont {P.}~\bibnamefont {Lazic}}, \bibinfo
  {author} {\bibfnamefont {G.}~\bibnamefont {Hautier}}, \bibinfo {author}
  {\bibfnamefont {K.}~\bibnamefont {Persson}},\ and\ \bibinfo {author}
  {\bibfnamefont {G.}~\bibnamefont {Ceder}},\ }\bibfield  {title} {\bibinfo
  {title} {First principles high throughput screening of oxynitrides for
  water-splitting photocatalysts},\ }\href {https://doi.org/10.1039/C2EE23482C}
  {\bibfield  {journal} {\bibinfo  {journal} {Energy Environ. Sci.}\ }\textbf
  {\bibinfo {volume} {6}},\ \bibinfo {pages} {157} (\bibinfo {year}
  {2013})}\BibitemShut {NoStop}%
\bibitem [{\citenamefont {Ishikawa}\ and\ \citenamefont
  {Miyake}(2020)}]{Ishikawa2020}%
  \BibitemOpen
  \bibfield  {author} {\bibinfo {author} {\bibfnamefont {T.}~\bibnamefont
  {Ishikawa}}\ and\ \bibinfo {author} {\bibfnamefont {T.}~\bibnamefont
  {Miyake}},\ }\bibfield  {title} {\bibinfo {title} {Evolutionary construction
  of a formation-energy convex hull: Practical scheme and application to a
  carbon-hydrogen binary system},\ }\href
  {https://doi.org/10.1103/PhysRevB.101.214106} {\bibfield  {journal} {\bibinfo
   {journal} {Phys. Rev. B}\ }\textbf {\bibinfo {volume} {101}},\ \bibinfo
  {pages} {214106} (\bibinfo {year} {2020})}\BibitemShut {NoStop}%
\bibitem [{\citenamefont {Ektarawong}\ \emph {et~al.}(2023)\citenamefont
  {Ektarawong}, \citenamefont {Johansson}, \citenamefont {Pakornchote},
  \citenamefont {Bovornratanaraks},\ and\ \citenamefont
  {Alling}}]{Ektarawong2023}%
  \BibitemOpen
  \bibfield  {author} {\bibinfo {author} {\bibfnamefont {A.}~\bibnamefont
  {Ektarawong}}, \bibinfo {author} {\bibfnamefont {E.}~\bibnamefont
  {Johansson}}, \bibinfo {author} {\bibfnamefont {T.}~\bibnamefont
  {Pakornchote}}, \bibinfo {author} {\bibfnamefont {T.}~\bibnamefont
  {Bovornratanaraks}},\ and\ \bibinfo {author} {\bibfnamefont {B.}~\bibnamefont
  {Alling}},\ }\bibfield  {title} {\bibinfo {title} {Boron vacancy-driven
  thermodynamic stabilization and improved mechanical properties of alb2-type
  tantalum diborides as revealed by first-principles calculations},\ }\href
  {https://doi.org/10.1088/2515-7639/acbe69} {\bibfield  {journal} {\bibinfo
  {journal} {Journal of Physics: Materials}\ }\textbf {\bibinfo {volume} {6}},\
  \bibinfo {pages} {025002} (\bibinfo {year} {2023})}\BibitemShut {NoStop}%
\bibitem [{\citenamefont {Kingma}\ \emph {et~al.}(2021)\citenamefont {Kingma},
  \citenamefont {Salimans}, \citenamefont {Poole},\ and\ \citenamefont
  {Ho}}]{Kingma2021vdm}%
  \BibitemOpen
  \bibfield  {author} {\bibinfo {author} {\bibfnamefont {D.~P.}\ \bibnamefont
  {Kingma}}, \bibinfo {author} {\bibfnamefont {T.}~\bibnamefont {Salimans}},
  \bibinfo {author} {\bibfnamefont {B.}~\bibnamefont {Poole}},\ and\ \bibinfo
  {author} {\bibfnamefont {J.}~\bibnamefont {Ho}},\ }\bibfield  {title}
  {\bibinfo {title} {On density estimation with diffusion models},\ }in\ \href
  {https://openreview.net/forum?id=2LdBqxc1Yv} {\emph {\bibinfo {booktitle}
  {Advances in Neural Information Processing Systems}}},\ \bibinfo {editor}
  {edited by\ \bibinfo {editor} {\bibfnamefont {A.}~\bibnamefont
  {Beygelzimer}}, \bibinfo {editor} {\bibfnamefont {Y.}~\bibnamefont
  {Dauphin}}, \bibinfo {editor} {\bibfnamefont {P.}~\bibnamefont {Liang}},\
  and\ \bibinfo {editor} {\bibfnamefont {J.~W.}\ \bibnamefont {Vaughan}}}\
  (\bibinfo {year} {2021})\BibitemShut {NoStop}%
\bibitem [{\citenamefont {Klicpera}\ \emph {et~al.}(2021)\citenamefont
  {Klicpera}, \citenamefont {Becker},\ and\ \citenamefont
  {G{\"u}nnemann}}]{Klicpera2021gemnet}%
  \BibitemOpen
  \bibfield  {author} {\bibinfo {author} {\bibfnamefont {J.}~\bibnamefont
  {Klicpera}}, \bibinfo {author} {\bibfnamefont {F.}~\bibnamefont {Becker}},\
  and\ \bibinfo {author} {\bibfnamefont {S.}~\bibnamefont {G{\"u}nnemann}},\
  }\bibfield  {title} {\bibinfo {title} {Gemnet: Universal directional graph
  neural networks for molecules},\ }in\ \href
  {https://openreview.net/forum?id=HS_sOaxS9K-} {\emph {\bibinfo {booktitle}
  {Advances in Neural Information Processing Systems}}},\ \bibinfo {editor}
  {edited by\ \bibinfo {editor} {\bibfnamefont {A.}~\bibnamefont
  {Beygelzimer}}, \bibinfo {editor} {\bibfnamefont {Y.}~\bibnamefont
  {Dauphin}}, \bibinfo {editor} {\bibfnamefont {P.}~\bibnamefont {Liang}},\
  and\ \bibinfo {editor} {\bibfnamefont {J.~W.}\ \bibnamefont {Vaughan}}}\
  (\bibinfo {year} {2021})\BibitemShut {NoStop}%
\bibitem [{\citenamefont {Joshi}\ \emph {et~al.}(2023)\citenamefont {Joshi},
  \citenamefont {Bodnar}, \citenamefont {Mathis}, \citenamefont {Cohen},\ and\
  \citenamefont {Lio}}]{Joshi2023}%
  \BibitemOpen
  \bibfield  {author} {\bibinfo {author} {\bibfnamefont {C.~K.}\ \bibnamefont
  {Joshi}}, \bibinfo {author} {\bibfnamefont {C.}~\bibnamefont {Bodnar}},
  \bibinfo {author} {\bibfnamefont {S.~V.}\ \bibnamefont {Mathis}}, \bibinfo
  {author} {\bibfnamefont {T.}~\bibnamefont {Cohen}},\ and\ \bibinfo {author}
  {\bibfnamefont {P.}~\bibnamefont {Lio}},\ }\href
  {https://openreview.net/forum?id=Rkxj1GXn9_} {\bibinfo {title} {On the
  expressive power of geometric graph neural networks}} (\bibinfo {year}
  {2023})\BibitemShut {NoStop}%
\bibitem [{\citenamefont {Kresse}\ and\ \citenamefont
  {Furthm\"{u}ller}(1996{\natexlab{a}})}]{Kresse1996-1}%
  \BibitemOpen
  \bibfield  {author} {\bibinfo {author} {\bibfnamefont {G.}~\bibnamefont
  {Kresse}}\ and\ \bibinfo {author} {\bibfnamefont {J.}~\bibnamefont
  {Furthm\"{u}ller}},\ }\href@noop {} {\bibfield  {journal} {\bibinfo
  {journal} {Computat. Mater. Sci.}\ }\textbf {\bibinfo {volume} {6}},\
  \bibinfo {pages} {15} (\bibinfo {year} {1996}{\natexlab{a}})}\BibitemShut
  {NoStop}%
\bibitem [{\citenamefont {Kresse}\ and\ \citenamefont
  {Furthm\"{u}ller}(1996{\natexlab{b}})}]{Kresse1996-2}%
  \BibitemOpen
  \bibfield  {author} {\bibinfo {author} {\bibfnamefont {G.}~\bibnamefont
  {Kresse}}\ and\ \bibinfo {author} {\bibfnamefont {J.}~\bibnamefont
  {Furthm\"{u}ller}},\ }\href@noop {} {\bibfield  {journal} {\bibinfo
  {journal} {Phys. Rev. B}\ }\textbf {\bibinfo {volume} {54}},\ \bibinfo
  {pages} {11169} (\bibinfo {year} {1996}{\natexlab{b}})}\BibitemShut {NoStop}%
\bibitem [{\citenamefont {Perdew}\ \emph {et~al.}(1996)\citenamefont {Perdew},
  \citenamefont {Burke},\ and\ \citenamefont {Ernzerhof}}]{Perdew1996}%
  \BibitemOpen
  \bibfield  {author} {\bibinfo {author} {\bibfnamefont {J.~P.}\ \bibnamefont
  {Perdew}}, \bibinfo {author} {\bibfnamefont {K.}~\bibnamefont {Burke}},\ and\
  \bibinfo {author} {\bibfnamefont {M.}~\bibnamefont {Ernzerhof}},\ }\href@noop
  {} {\bibfield  {journal} {\bibinfo  {journal} {Phys. Rev. Lett.}\ }\textbf
  {\bibinfo {volume} {77}},\ \bibinfo {pages} {3865} (\bibinfo {year}
  {1996})}\BibitemShut {NoStop}%
\bibitem [{\citenamefont {Bl\"{o}chl}(1994)}]{Blochl1994}%
  \BibitemOpen
  \bibfield  {author} {\bibinfo {author} {\bibfnamefont {P.~E.}\ \bibnamefont
  {Bl\"{o}chl}},\ }\href@noop {} {\bibfield  {journal} {\bibinfo  {journal}
  {Phys. Rev. B}\ }\textbf {\bibinfo {volume} {50}},\ \bibinfo {pages} {17953}
  (\bibinfo {year} {1994})}\BibitemShut {NoStop}%
\bibitem [{\citenamefont {Monkhorst}\ and\ \citenamefont
  {Pack}(1976)}]{Monkhorst1976}%
  \BibitemOpen
  \bibfield  {author} {\bibinfo {author} {\bibfnamefont {H.~J.}\ \bibnamefont
  {Monkhorst}}\ and\ \bibinfo {author} {\bibfnamefont {J.~D.}\ \bibnamefont
  {Pack}},\ }\href@noop {} {\bibfield  {journal} {\bibinfo  {journal} {Phys.
  Rev. B}\ }\textbf {\bibinfo {volume} {13}},\ \bibinfo {pages} {5188}
  (\bibinfo {year} {1976})}\BibitemShut {NoStop}%
\bibitem [{\citenamefont {Pack}\ and\ \citenamefont
  {Monkhorst}(1977)}]{Pack1977}%
  \BibitemOpen
  \bibfield  {author} {\bibinfo {author} {\bibfnamefont {J.~D.}\ \bibnamefont
  {Pack}}\ and\ \bibinfo {author} {\bibfnamefont {H.~J.}\ \bibnamefont
  {Monkhorst}},\ }\href@noop {} {\bibfield  {journal} {\bibinfo  {journal}
  {Phys. Rev. B}\ }\textbf {\bibinfo {volume} {16}},\ \bibinfo {pages} {1748}
  (\bibinfo {year} {1977})}\BibitemShut {NoStop}%
\bibitem [{\citenamefont {Gasteiger}\ \emph {et~al.}(2020)\citenamefont
  {Gasteiger}, \citenamefont {Groß},\ and\ \citenamefont
  {Günnemann}}]{Gasteiger2020dimenet}%
  \BibitemOpen
  \bibfield  {author} {\bibinfo {author} {\bibfnamefont {J.}~\bibnamefont
  {Gasteiger}}, \bibinfo {author} {\bibfnamefont {J.}~\bibnamefont {Groß}},\
  and\ \bibinfo {author} {\bibfnamefont {S.}~\bibnamefont {Günnemann}},\
  }\bibfield  {title} {\bibinfo {title} {Directional message passing for
  molecular graphs},\ }in\ \href {https://openreview.net/forum?id=B1eWbxStPH}
  {\emph {\bibinfo {booktitle} {International Conference on Learning
  Representations}}}\ (\bibinfo {year} {2020})\BibitemShut {NoStop}%
\end{thebibliography}%

\pagebreak
\newpage

\widetext
\title{Diffusion probabilistic models combined with variational autoencoder for crystal structure generation}
\begin{center}
\textbf{\large Supplementary Materials}
\end{center}

\newcommand{\beginsupplement}{%
        \setcounter{table}{0}
        \renewcommand{\thetable}{S\arabic{table}}%
        \setcounter{figure}{0}
        \renewcommand{\thefigure}{S\arabic{figure}}%
	\renewcommand{\theHfigure}{S\arabic{figure}}
        \setcounter{equation}{0}
        \renewcommand{\theequation}{S\arabic{equation}}%
        \setcounter{section}{0}
        \renewcommand{\thesection}
        {S\arabic{section}}%
     }
     
\beginsupplement
\section{Encoders}
\begin{figure}[h]
    \centering    \includegraphics[width=0.95\textwidth]{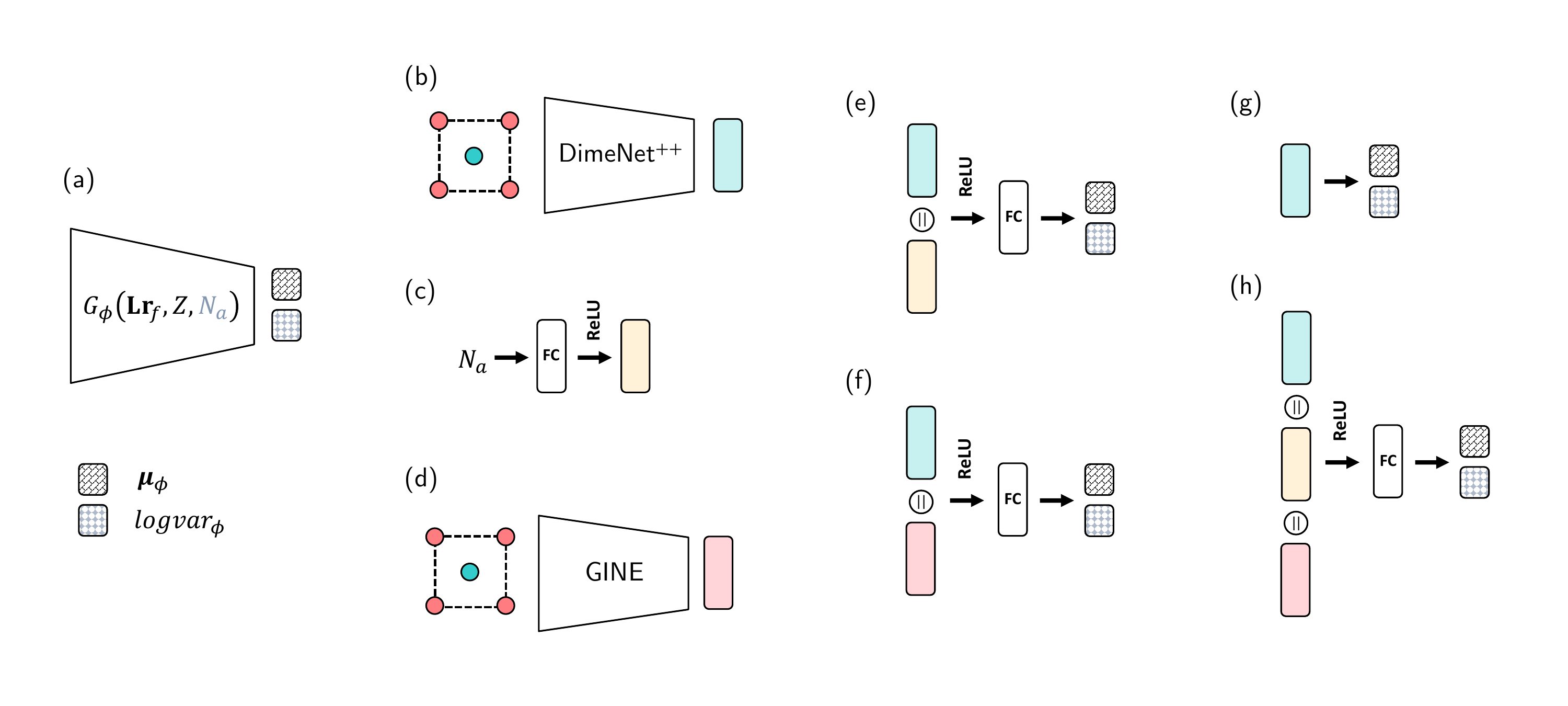}
    \caption{(a) is the encoder for predicting $\boldsymbol{\mu}_{\phi}$ and $logvar_{\phi}$. (b) and (d) are DimeNet$^{++}$ and GINE encoders, respectively, that take pristine crystal structures as inputs. (c) is the $N_a$ encoder that takes the number of atoms $N_a$ as an input. (e), (f), (g), and (h) are encoders of DP-CDVAE+$N_a$, DP-CDVAE+GINE, DP-CDVAE, and DP-CDVAE+$N_a$+GINE models, respectively. Blue, yellow, and pink boxes are the latent features from (b), (c), and (d), respectively, and FC boxes are fully connected layers.}
    \label{fig:encoders}
\end{figure}

\section{Model evaluation}
\begin{table*}[h]
\caption{Reconstruction performance using Eq.~\ref{eq:r_t-1}.}
\label{tab:si_reconstruction}
\begin{ruledtabular}
\begin{tabular}{lcccccc}
\multicolumn{1}{l}{Models} & \multicolumn{3}{c}{Match rate (\%) $\uparrow$} & \multicolumn{3}{c}{$\langle\delta_{\text{rms}}\rangle$ $\downarrow$} \\
\cline{2-4} \cline{5-7}
 & Perov-5 & Carbon-24 & MP-20 & Perov-5 & Carbon-24 & MP-20 \\
\hline
DP-CDVAE & 68.72 & 35.86 & 20.40 & 0.0177 & 0.2573 & 0.0715 \\
DP-CDVAE+$N_a$ & 67.56 & 38.13 & 21.71 & 0.0222 & 0.2719 & 0.0774 \\
DP-CDVAE+GINE & 49.56 & 35.27 & 19.98 & 0.0822 & 0.2683 & 0.0620 \\
DP-CDVAE+$N_a$+GINE & 64.12 & 34.68 & 24.43 & 0.0250 & 0.3212 & 0.0697 \\
\end{tabular}
\end{ruledtabular}
\end{table*}

\begin{figure*}[h]
\includegraphics[width=0.98\textwidth]{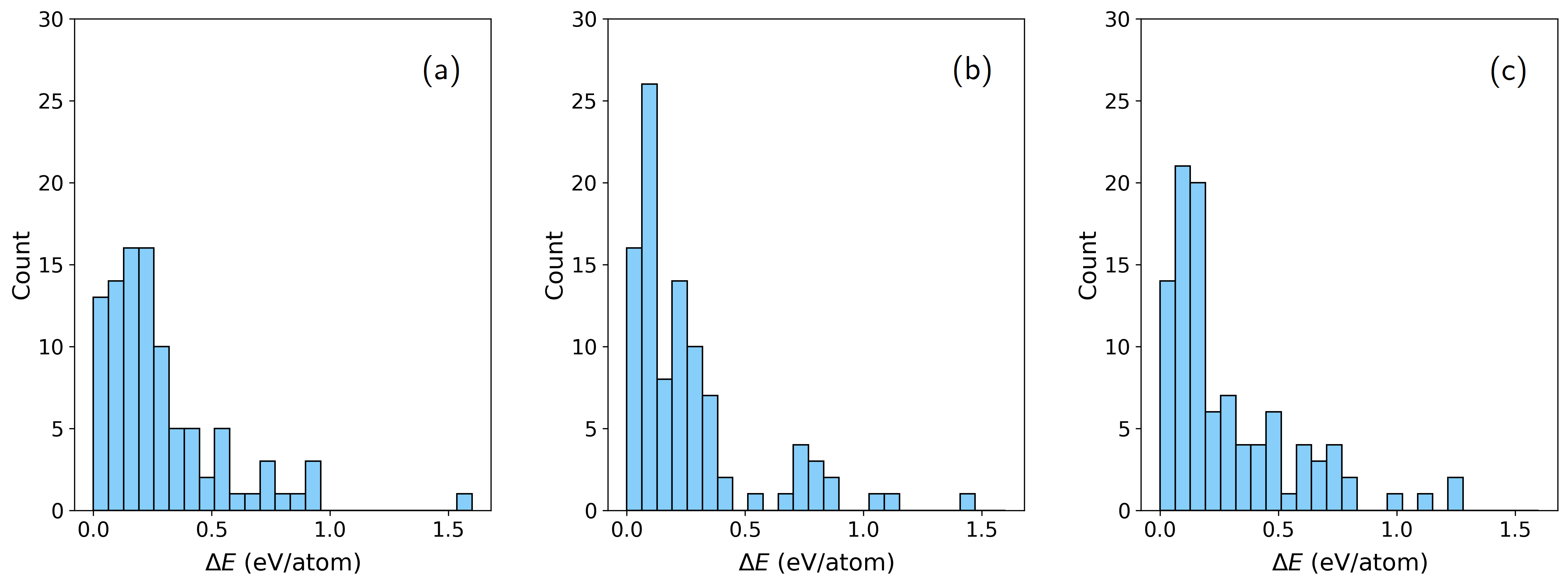}
\caption{\label{fig:histograms}
Histograms of energy difference between generated and relaxed structures of (a) CDVAE, (b) CDVAE+Fourier, and (c) DP-CDVAE models. The modes (highest counting number) of CDVAE, CDVAE+Fourier, and DP-CDVAE models are 19.2 -- 32.0, 68.0 -- 128, and 64.0 -- 128 eV/atom, respectively, where the bin size is 64.0 meV/atom. 
}
\end{figure*}
\newpage
\section{Fourier features}
\label{sec:fourier}
We appended the Fourier features to the node attributes for the input of the diffusion network. The Fourier features are the concatenation of $\sin(2^n\pi\mathbf{r}_t)$ and $\cos(2^n\pi\mathbf{r}_t)$ where $n \in \{n_{\text{min}}, ..., n_{\text{max}}\}$. For every DP-CDVAE model in the main text, $n_{\text{min}}=3$ and $n_{\text{max}}=8$. 

\section{Loss Functions}
Similar to the original CDVAE, the total loss function to train the model consists of 5 sub-loss functions: the Kullback–Leibler divergence loss ($\mathcal{L}_{KLD}$) for training $\boldsymbol{\mu}_{\phi}$ and $logvar_{\phi}$, the lattice loss ($\mathcal{L}_{latt}$) for training lattice parameters, the composition loss ($\mathcal{L}_{comp}$) for training $\mathbf{A}_{\mathbf{z}}$, the loss for training the number of atoms ($\mathcal{L}_{N_a}$), and the loss from the diffusion network ($\mathcal{L}_{diff}$) for training $\boldsymbol{\epsilon}_{\theta}$ and $\mathbf{A}_{\theta}$. In particular, the total loss is
\begin{equation} \label{eq:total_loss}
    \mathcal{L} = \mathcal{L}_{diff} + \lambda_1 \mathcal{L}_{KLD} + \lambda_2 \mathcal{L}_{latt} + \lambda_3 \mathcal{L}_{comp} + \lambda_4 \mathcal{L}_{N_a},
\end{equation}
where $\lambda_1$, $\lambda_2$, $\lambda_3$, and $\lambda_4$ are tunable loss scaling factors, $\mathcal{L}_{latt}$ is computed from the mean square error of lattice parameters, and $\mathcal{L}_{comp}$ and $\mathcal{L}_{N_a}$ are cross-entropy losses.
\end{document}